\documentclass{article}

\PassOptionsToPackage{numbers,sort&compress}{natbib}
\usepackage[eandd,final]{neurips_2026}

\usepackage[utf8]{inputenc}
\usepackage[T1]{fontenc}
\usepackage{hyperref}
\usepackage{url}
\usepackage{booktabs}
\usepackage{amsfonts}
\usepackage{nicefrac}
\usepackage{microtype}
\usepackage{xcolor}
\usepackage{graphicx}
\usepackage{amsmath}
\usepackage{amssymb}
\graphicspath{{./}}
\usepackage{enumitem}

\title{The Complexity Kink: LLM Rubric Instruments for Causal Inference on Code Generation Reliability}

\newcommand{\codeurl}{\url{https://github.com/uwm-se/ComplexityKink}}
\newcommand{\dataurl}{\url{https://huggingface.co/datasets/uwm-software-engineering/complexity-kink-research}}

\author{%
  Michael Hernandez \\
  University of Wisconsin--Milwaukee \\
  \texttt{herna273@uwm.edu} \\
  \And
  Tian Zhao \\
  University of Wisconsin--Milwaukee \\
  \texttt{tzhao@uwm.edu}
}

\begin{document}

\maketitle

\begin{abstract}
Measuring complexity from a model's own code is biased by failure: a hard prompt can produce a short, broken program that looks simple. We instead score each prompt's structure on six dimensions from the prompt alone, never from generated code or test results. We draw 5{,}000 Python prompts across six bands of a preliminary single-rater score, and four out-of-panel LLM judges rescore every prompt (19{,}997 scores; composite ICC$(2,1)=0.872$ on the 4{,}998 prompts with all four). Twenty-one models answer every prompt (105{,}000 generations), and an independent LLM auditor, checked against known-answer cases, reviews each unit-test verdict. Averaged over models, pass rate has a nonmonotone breakpoint at composite $\hat{\gamma}=14.0$: 83.4\% at or below, 90.1\% above. That jump is not a law of difficulty. With task-type fixed effects the breakpoint stays at 14.0 but the adjusted jump shrinks from 6.8 to 0.4 points, and neither of the benchmark's two prompt sources reproduces the jump on its own. Nineteen models pass more often above their own breakpoint and two less often, and one of those two declines, Gemini 3.1 Pro Preview's, comes from 156 API calls that returned nothing. In post hoc follow-up, Kimi K2.5 shows the clearest decline: with a prompt-source control, and within the later source alone, its break is at 14.25, where its later-source pass rate falls from 0.848 to 0.808 (95\% interval $[14.00,14.75]$). A 365-prompt audit-clean extension matches the original raw test outcomes at bins 15 and 16 but adds only 14 prompts above bin 16. Among failed generations with measurable output complexity, 25.0\% pair a prompt composite above 8 with output complexity of at most 10, twice the share among passing ones. The pooled breakpoint moves to 11.25 on raw test outcomes and 13.75 under the partial audit used at submission. Human agreement with the index is moderate and rater-dependent, paraphrase and cross-language rescoring preserve its ordering, and overidentification tests reject the six dimensions as instruments, so we treat the composite as an index and make no causal claim. The contribution is a prompt-side way to locate where each model's pass rate changes and to test how much that depends on the prompt mix and on how success is scored.
\end{abstract}

\section{Introduction}
% =====================================================================

Code-generating LLMs are evaluated functionally: does the program pass its tests~\cite{chen2021evaluating,austin2021program,jimenez2023swebench}? That tells us how often a model succeeds, not how success changes with the structure a task demands. Two models with the same average pass rate can behave very differently on prompts that need interacting branches, data structures, edge cases, or multi-step algorithms.

The usual way to measure a task's complexity is after the fact, from the code the model wrote. But when a model fails, what it writes is often a short stub or a broken fragment with low measured cyclomatic complexity, so a demanding prompt can land in a low-complexity bin precisely because the model failed. That can flatten, shift, or hide the link between task structure and pass rate. We call it the \emph{Reverse Threshold Problem}: some of the ``simple'' code that fails is really hard tasks the model never finished. Counting keywords in the instruction avoids the problem but reduces structure to surface words. We instead ask a rubric six separate questions about branching, iteration, state, data structures, edge cases, and composition. It is still a proxy, but it looks only at the prompt, so a model's failure cannot leak into it.

We build the Complexity Kink benchmark around that separation. From OpenCodeInstruct~\cite{opencodeinst} we draw 5{,}000 Python prompts stratified on a preliminary single-rater score; four out-of-panel LLM judges rescore the locked prompts, 21 models each answer every prompt, and an independent auditor reviews every unit-test verdict. We then estimate where pass rate changes along the prompt score, for each model and for all models pooled, and ask how much each estimate depends on the model or pooling rule, the prompt source, the score, the generation settings, and the test protocol. We also report output-side comparisons and the instrumental-variable design behind the title, but its candidate instruments fail their joint exclusion restrictions, so we give them no causal reading.

The results do not support a universal cutoff. Averaged over models, pass rate falls through the middle of the index and rebounds in the high region, but task type and prompt source explain most of the jump: the two sources differ sharply in score and pass rate, and neither reproduces the pooled $+6.8$-point change. Nineteen model-specific fits are higher above their breakpoint and two lower, and one of the two comes from failed API calls. Kimi K2.5 gives the clearest decline, and it survives a prompt-source control and every outcome definition we tried. A matched extension agrees with the original data at bins 15 and 16, where it adds support, but the region above 16 is still too thin for a firm claim.

\vspace{2pt}
\noindent\textbf{Contributions.}
\begin{itemize}
\item We formalize failure-dependent output complexity and show it in this sample: 25.0\% of failed complete cases pair a structurally nontrivial prompt with generated-output cyclomatic complexity of at most 10.
\item We introduce an ensemble-scored prompt-side index and test it against human ratings, plain-language paraphrases, and Java and C++ re-expressions.
\item We release a 5{,}000-prompt Python benchmark with all judge scores, 105{,}000 generations, test outcomes, and audit verdicts,\footnote{Code and data: \codeurl{}.} and use it to estimate model-specific reliability breakpoints.
\item We stress-test those estimates under pooling, repeated sampling, task type, prompt source, outcome definition, and an audit-clean extension, and report when a breakpoint is composition-sensitive.
\end{itemize}

% =====================================================================
\section{Related Work}
% =====================================================================

\textbf{LLM code evaluation.} HumanEval~\cite{chen2021evaluating}, MBPP~\cite{austin2021program}, and SWE-Bench~\cite{jimenez2023swebench} established functional testing as the central protocol for code-generating LLMs; that functional outcome remains our target variable. We instead study how to measure the structure of the task being attempted. OpenCodeInstruct~\cite{opencodeinst} supplies instruction-code pairs, but reference code reflects one implementation rather than latent task structure.

\textbf{Complexity measurement and benchmark bias.} Cyclomatic complexity is computable from source code, from McCabe's formulation to static analyzers such as Lizard~\cite{mccabe1976complexity,lizard}. In code-generation evaluation the available source is often the model output, which makes the metric post-treatment: measured complexity depends on success or failure~\cite{montgomery2018posttreatment}. We score the prompt instead.

\textbf{Instrumental variables and threshold estimation.} IVs are a standard response to endogenous regressors~\cite{wooldridge2010econometric}, with weak-instrument and overidentification tests as essential diagnostics~\cite{stock2005testing,sargan1958estimation,hansen1982large,bowsher2002reject}, and are increasingly used in empirical software engineering~\cite{grafvlachy2024cleaning,siebert2023applications}. Our rubric-dimension IV fits fail these diagnostics and are kept only as diagnostics; the primary analysis applies Hansen's threshold regression and bootstrap sup-Wald test~\cite{hansen2000sample} to the composite directly.

\textbf{LLMs as judges.} G-Eval~\cite{liu2023geval}, AlpacaEval~\cite{alpacaeval}, and MT-Bench~\cite{zheng2023judging} use LLMs to evaluate model outputs, building on rubric-based automatic evaluation in NLP~\cite{hashimoto2019unifying}. Our rubric judges score the input prompt, not an answer, against a published structural rubric, and they sit outside the evaluated panel: they measure a property of the input rather than grade an output. A separate LLM auditor reviews unit-test verdicts (Section~\ref{sec:data}).

% =====================================================================
\section{Background: Failure-Dependent Output Complexity}
% =====================================================================

Let $\kappa_i^*$ denote the latent structure required by task $i$, let $y_{im}$ denote the functional outcome for model $m$, and let $\kappa_{im}^{\mathrm{obs}}$ be the cyclomatic complexity measured on that model's generated output. Even a passing output is only one implementation of the task, so $\kappa_{im}^{\mathrm{obs}}$ need not equal $\kappa_i^*$. More importantly, the measurement itself depends on the outcome:
\begin{equation}
\kappa_{im}^{\mathrm{obs}} = g(\kappa_i^*, y_{im}, \eta_{im}),
\label{eq:failure-dependent}
\end{equation}
where failed generations often collapse to short programs with low measured complexity. Stratifying pass rate by $\kappa_{im}^{\mathrm{obs}}$ therefore conditions on a post-generation quantity affected by success.

We avoid this with a prompt-side index $C_i$ that depends only on the prompt, so it cannot depend on whether a model succeeded. (Our judges scored the prompts after most generations existed, but never saw them.) Our primary target is descriptive: the conditional relationship $\mathbb{E}[y_i\mid C_i]$ in the constructed benchmark, where $y_i$ is pass rate averaged over evaluated models. The six rubric dimensions could also serve as candidate instruments for $\kappa_i^{\mathrm{obs}}$, but a causal reading would need strong exclusion restrictions, and the diagnostics below reject them.
% =====================================================================
\section{Methodology}
% =====================================================================

\subsection{Data}\label{sec:data}
We use 5{,}000 Python tasks from OpenCodeInstruct~\cite{opencodeinst} from two sources, which we call construction frames: 2{,}246 prompts kept from an earlier stage, drawn from a 200{,}000-record source-ordered prefix, and 2{,}754 later candidates that passed automated contract and test-quality filtering and deliberately added high reference-complexity tasks. The earlier prompts never went through that filter: for 1{,}259 of them the source's own reference solution fails at least one unit test. A preliminary \texttt{o4-mini} rater scored candidates with the rubric below, and we sampled 834 prompts in each of the preliminary-score bands 0--3 and 4--6 and 833 in each of 7--9, 10--12, 13--15, and 16--24. That score was used for sampling only.

After the prompt set was locked, the four-judge ensemble rescored every prompt, from the prompt text alone, and every analysis uses that ensemble mean. On the bands $[0,3]$, $(3,6]$, $(6,9]$, $(9,12]$, $(12,15]$, $(15,24]$ the final index has 319, 943, 940, 898, 1{,}425, and 475 prompts: stratified by the preliminary score but not balanced on the final one, and unlike the natural OpenCodeInstruct distribution. Reference-solution cyclomatic complexity shaped the upstream pool and is used only for alignment checks.

The evaluated panel has 21 models: Claude Opus 4.6, Opus 4.7, and Sonnet 4.6; GPT-5.4, GPT-5-mini, GPT-4.1, GPT-OSS-20B, and GPT-OSS-120B; Gemini 3.1 Pro Preview and Gemini 3 Flash; Grok-3; DeepSeek V3.2; Kimi K2.5; Qwen 3.6 Plus and Qwen 3.5-9B; Mistral Large-3, Devstral Small 2505, and Ministral-3-14B-reasoning; Llama 3.3-70B; GLM 4.7-flash; and Trinity-large. No rubric judge is in the panel. Five open-weight models were served locally as 4- or 8-bit quantizations, and each model's two frames were generated in separate runs, weeks apart and sometimes through different routes or token limits; the release records routes, versions, and settings. A locally served, quantized AuroraGPT-IT-v4 run was generated only on an earlier-stage 5{,}000-prompt set, which includes the 2{,}246 retained prompts but none of the 2{,}754 later ones, and was excluded post hoc without a prespecified rule; under the submission audit (defined below) its pass rate was 20.6\% on that set and 23.2\% on the 2{,}246 retained prompts, and performance was not a documented exclusion criterion. Each model generated one solution per prompt ($21\times5{,}000=105{,}000$ generations).

Pass rate starts from the fraction of Nvidia-supplied unit tests passed. Because tests can reject valid code (say, a renamed function) or miss defects, MiMo-V2.6-Pro, an auditor from a vendor used nowhere else in the study, read each generation's task, code, tests, and harness result. A ``correct'' verdict sets pass rate to 1.0 and ``incorrect'' to 0.0; uncertain, unparseable, or cut-off responses (1.1\%) keep the harness value, and the 208 generations with no code count as failures. The outcome is therefore a binary verdict for 98.9\% of generations; mean pass is 0.851, against 0.792 on raw harness values. Appendix~\ref{app:outcome-audit} covers the known-answer checks, where selection departed from our protocol, second-auditor agreement, and the main analyses on raw harness outcomes and under the partial \texttt{o4-mini} audit, of the earlier frame only, that the submitted version of this paper used (the \emph{submission audit}). The pooled analysis averages pass rate over the 21 models within prompt ($N=5{,}000$); per-model analyses use the same 5{,}000 rows. Generated-output complexity is McCabe cyclomatic complexity from Lizard~\cite{lizard}, summed over the functions in the cleaned code; it exists for 103{,}948 generations, and analyses using it report complete-case denominators.

\subsection{Rubric Scoring}

Four out-of-panel LLM judges deployed through Azure AI Foundry, \texttt{o4-mini}, \texttt{gpt-5.5}, \texttt{llama-4-maverick}, and \texttt{command-a}, score every prompt with the same rubric prompt: 4{,}998 prompts have four scores, one has three, and one has two (19{,}997 score rows). We average each prompt-dimension pair across available judges; composite ICC$(2,1)$ (two-way random effects, absolute agreement, single rater) is 0.872 on the 4{,}998 complete cases, and scorer disagreement is kept for diagnostics. Each prompt is scored on six dimensions, each $0$ to $4$ (Table~\ref{tab:complexity_dimensions}).
\begin{table}[h]
\caption{Complexity dimensions and scoring criteria.}
\label{tab:complexity_dimensions}
\centering
\begin{tabular}{p{3cm} p{4cm} p{5.5cm}}
\hline
\textbf{Dimension} & \textbf{Score = 0} & \textbf{Score = 4} \\
\hline
Branching & No conditionals & Deeply nested or combinatorial \\
Iteration & None & Nested recursion or complex multi-pass \\
State & Stateless or single variable & Concurrent state tracking \\
Data structures & Primitives only & Multiple interacting complex structures \\
Edge cases & None & Combinatorial interacting checks \\
Composition & Single operation & Pipeline of interdependent algorithms \\
\hline
\end{tabular}
\end{table}

The primary prompt-side index is $C_i = \sum_{d=1}^{6} Z_{i,d} \in [0,24]$, where $Z_{i,d}$ is the ensemble mean for dimension $d$. The dimension vector $\mathbf{Z}_i=(Z_{i,1},\ldots,Z_{i,6})$ is kept for interpretation, reliability analysis, and secondary IV diagnostics.

\textbf{Design decisions.} Separate dimensions make the index interpretable, show where raters disagree, allow per-dimension human calibration, and permit an overidentification test when treated as candidate IVs. The five-point scale leaves a neutral middle score for ambiguous prompts. Each dimension is anchored to expected solution structure and explicitly excludes LLM difficulty, code length, and language idioms, so the outcome is not built into the measure. Keeping the judges out of the evaluated panel reduces circularity but does not by itself establish accuracy, human agreement, or IV exclusion.

\subsection{Primary Breakpoint Analysis}\label{sec:breakpoint}
For a candidate breakpoint $\gamma$, we fit separate linear relationships on the two sides of the prompt-side composite:
\begin{equation}
\mathbb{E}[y_i\mid C_i] =
\begin{cases}
\alpha_1+\beta_1 C_i, & C_i\leq\gamma,\\
\alpha_2+\beta_2 C_i, & C_i>\gamma.
\end{cases}
\label{eq:threshold}
\end{equation}
The mean-pooled outcome averages pass rate over the 21 models within prompt ($N=5{,}000$); model-specific fits use the same index and tests, and we also report median pooling and 21 leave-one-model-out fits. Regime means (averages on each side of the break) describe a selection-stratified sample. A model's $\hat{\gamma}$ marks where its pass-rate curve changes on this benchmark; it is not a causal, universal, or deployment-level failure point, and we read direction from the average pass rate on each side, not from the test statistic.

To locate the breakpoint, we adapt Hansen's threshold-regression procedure~\cite{hansen2000sample}:
\begin{enumerate}
\item For each candidate $\gamma$ on a percentile-spaced grid over $C_i$, requiring at least 500 prompts on each side, we compare (\ref{eq:threshold}) with a single line using a Wald-type $F$-statistic $W(\gamma)$.
\item The supremum $W^* = \sup_\gamma W(\gamma)$ identifies the threshold $\hat{\gamma}$.
\item Under $H_0$ of no threshold, we use Rademacher wild-bootstrap residual resampling and recompute $W^*$ at each draw. A separate pairs bootstrap gives a percentile interval for $\hat{\gamma}$, which we treat as descriptive.
\end{enumerate}
We repeat the pooled search with nine task-type indicators and, separately, the frame indicator, in two ways: as ordinary additive fixed effects, where only the composite line may break, and, as in the submitted analysis, letting the indicators' effects also differ on each side. For each we report the adjusted jump, the coefficient on $\mathbf{1}[C_i>\hat{\gamma}]$ in a regression on that indicator and the controls. We also compare frame-specific linear slopes and repeat the search within each frame, for the pooled fit and, post hoc, for models whose pass rate falls across their break. Bootstrap and placebo settings are in Appendix~\ref{app:robustness}.

\subsection{Output-Complexity and IV Diagnostics}\label{sec:hansen-j}
We first compare pass rate directly with generated-output complexity and count the reverse-threshold cell among failed generations. As a secondary diagnostic, we regress generated-output complexity on the six rubric dimensions and use the fitted value in a 2SLS pass-rate regression:
\begin{align}
\kappa_i^{\mathrm{obs}} &= \pi_0+\mathbf{Z}_i\boldsymbol{\pi}+\nu_i, \label{eq:stage1}\\
y_i &= \alpha+\beta\widehat{\kappa}_i+u_i. \label{eq:stage2}
\end{align}
We estimate these with \texttt{IV2SLS} from \texttt{linearmodels}~\cite{linearmodels} and report Sargan (five degrees of freedom) and robust Wooldridge overidentification tests, single-instrument fits, principal-component rotations, and post hoc subsets. These tests reject the full instrument set, so $\hat{\beta}$ is a diagnostic coefficient, not a causal estimate.

\subsection{Additional Robustness Designs}
After the initial analysis we added post hoc stress tests, none preregistered: a nine-category task taxonomy, the frame analyses above, a five-draw repeated-sampling subset, plain-language paraphrase and Java/C++ re-expression rescoring, a blind two-grader human calibration, a 365-prompt audit-clean high-complexity extension, and a reanalysis that treats generations with no API response as missing. Appendix~\ref{app:robustness-designs} details the designs, and Figure~\ref{fig:pipeline} there summarizes the workflow.

% =====================================================================
\section{Results}
% =====================================================================

\subsection{Alignment with Reference Complexity}

Reference-solution complexity helped build the candidate pool, so agreement with it is a sanity check, not independent validation. On all 5{,}000 prompts, the composite has Pearson $r=0.80$ and Spearman $\rho=0.81$ with reference $\kappa$, rising from 4.64 at $\kappa=1$ to 14.34 at $\kappa\in[21,50]$, though not at every adjacent bin (Appendix Table~\ref{tab:rubric-ref-cc}). At the low end, where failed stubs push output CC toward 1, the rubric separates $\kappa=1$ from $\kappa=2$ prompts ($+0.61$, $p=3.1\times10^{-4}$), driven by iteration, state, and branching. A lexical keyword baseline also separates them ($+0.58$, $p=2.4\times10^{-3}$; Appendix Table~\ref{tab:low-end-separation}), so the rubric does not beat every cheap proxy; its advantages are explicit structural dimensions, measurable rater uncertainty, and the calibration checks below.

\subsection{Rater and Measurement Calibration}\label{sec:calibration}

Beyond the four-judge ICC of 0.872, we ran a deliberately difficult human calibration with LLM scores hidden: the first author graded 200 prompts, roughly half sampled for high judge disagreement, and the second author graded 50 of them. On all 200, first-author agreement with the ensemble is Pearson $r=0.408$, Spearman $\rho=0.373$, ICC$(2,1)=0.395$, with mean offset $-1.18$ points; per-dimension Pearson correlations range from 0.249 to 0.484 and quadratic-weighted $\kappa$ from 0.217 to 0.435. On the shared 50, the graders correlate 0.561 with each other and 0.613 and 0.921 with the ensemble, with offsets $-0.69$ and $-1.35$. These are stress-test estimates with substantial rater variation, not a human ground truth.

Two checks ask whether the index follows presentation rather than structure. Across 150 plain-language rewrites, ranks barely move (Spearman $\rho=0.963$); the mean shift is small ($-0.115$) but detectable (paired $p=0.031$), and 91.3\% of pairs are within one point. The rewrites were not verified by execution, so this shows score stability under a preservation instruction, not semantic equivalence. For 117 prompts re-expressed in Java and C++, the composite correlates 0.992 and 0.969 with Python, with inter-judge ICC 0.912 (Python), 0.913 (Java), and 0.882 (C++): evidence that scores transfer, not that non-Python generation is reliable.

\subsection{Generated-Output Complexity Misses the Prompt-Side Pattern}

\begin{figure}[h]
\centering
\includegraphics[width=\linewidth]{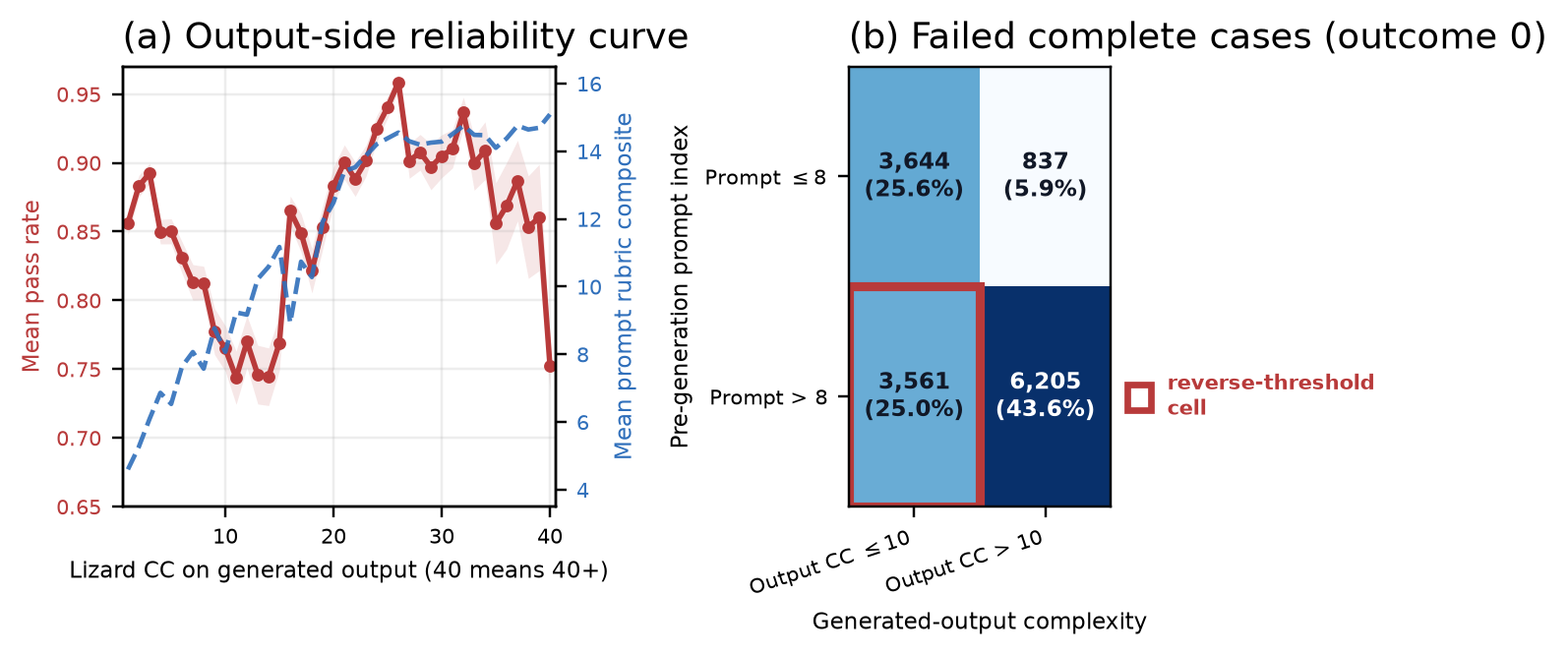}
\caption{Generated-output complexity is failure-dependent. (a) Across 103{,}948 complete-case generations, points show mean pass rate by integer Lizard output CC; the shaded descriptive band is $\pm1.96$ generation-row standard errors and does not adjust for repeated prompts or models. Here 40 denotes CC $\geq40$. The dashed blue line is the conditional mean prompt composite, read on the right axis. (b) Among failed complete cases (outcome 0), the outlined cell contains 3{,}561 of 14{,}247 generations (25.0\%): the prompt composite is $>8$ while generated-output CC is $\leq10$.}
\label{fig:output-cc}
\end{figure}

Figure~\ref{fig:output-cc} plots pass rate against the Lizard complexity of the generated code: the curve is nonmonotone, with no clean decline. Among the 14{,}247 failed generations (outcome 0) with a computable Lizard value, 3{,}561 (25.0\%) have prompt composite $>8$ but output CC $\leq10$, twice the share among passing generations (12.2\%). Because 1.1\% of outcomes keep their harness value, defining ``failed'' as pass rate at most 0.25, 0.50, or 0.65 barely changes the share (25.0\%, 25.2\%, 25.4\%). An automated pipeline would quietly file these demanding prompts as simple. Another 969 failed generations have no computable output CC and are left out.

Among passing generations only, where output CC is less contaminated by failed stubs, a descriptive linear mapping has $R^2=0.614$ and maps prompt $\hat{\gamma}=14.0$ to output CC of about 23.6. This is a scale translation within a selected sample, not a second threshold estimate. The OLS and 2SLS comparisons are in Appendix~\ref{app:econ-diagnostics}.

\subsection{The Mean-Pooled Breakpoint}\label{sec:pooled}

Averaged over models, the sup-Wald search selects $\hat{\gamma}=14.0$ on the prompt composite (sup-Wald $=151.15$). None of 2{,}000 wild-bootstrap draws or 2{,}000 shuffled-index placebo draws reaches it ($p_{\mathrm{MC}}\leq1/2001<0.001$ each). The 3{,}703 prompts at or below $\hat{\gamma}$ average 83.4\% pass; the 1{,}297 above average 90.1\%. This is not a monotone collapse: pass rate falls through the middle of the index to a trough near composite 10, rebounds to a peak at bin 15, and falls again in the sparse bins above. By BIC the piecewise form beats linear and cubic fits ($-67$ versus 209 and 33).

\begin{figure}[h]
\centering
\includegraphics[width=\linewidth]{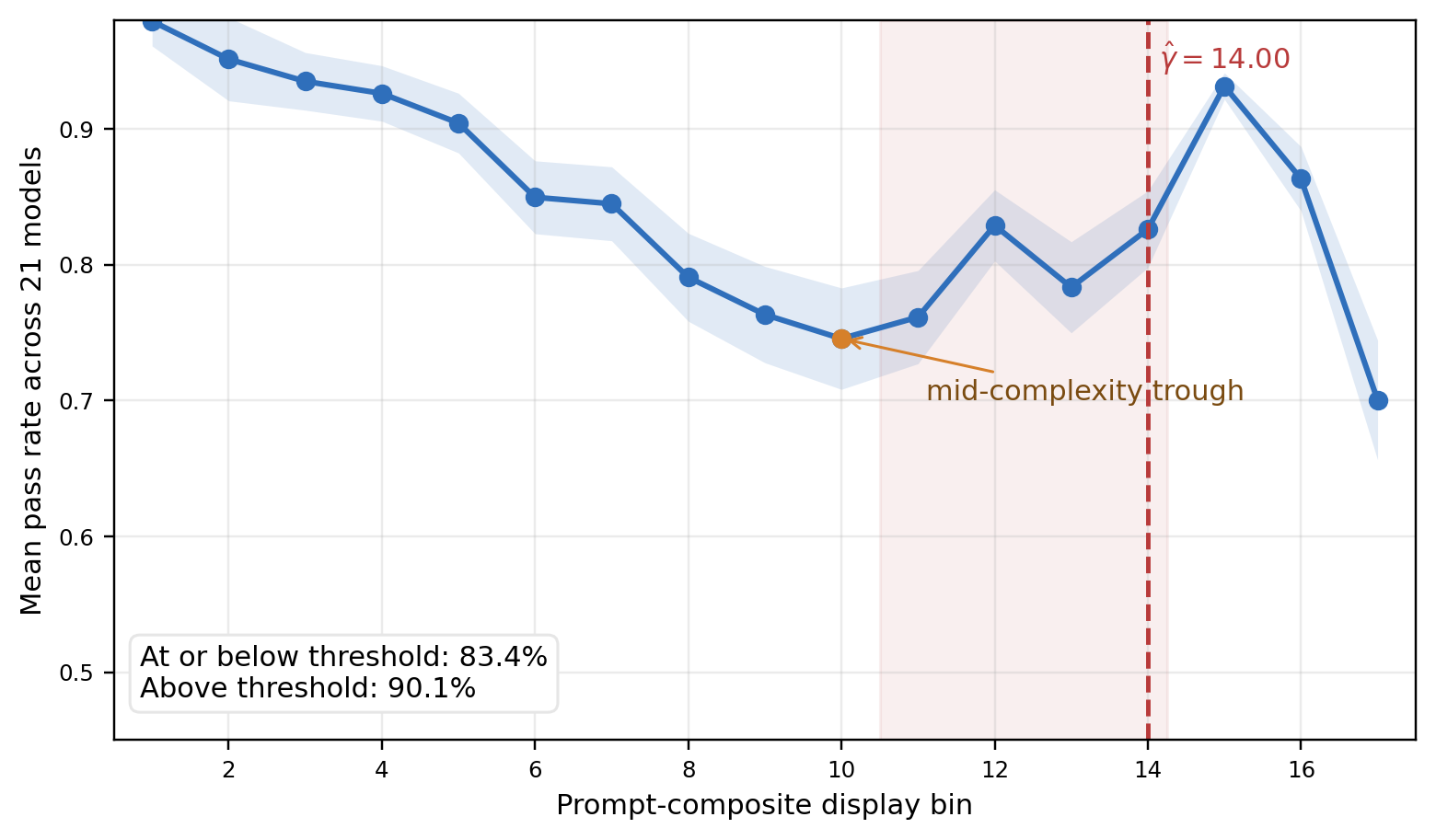}
\caption{Unadjusted mean-pooled relationship in the original 5{,}000-prompt benchmark. Pass rate is first averaged across 21 models within prompt. Display bin $b$ contains $C_i\in[b-0.5,b+0.5)$, so exact half-point boundaries enter the higher bin. The dashed line marks the threshold $\hat{\gamma}=14.0$ estimated from the unbinned composite; the shaded vertical band is the 95\% pairs-bootstrap interval. Bin 17 contains only 45 prompts and is descriptive; bins 0, 18, and 19 (3, 6, and 5 prompts) are omitted. This figure does not include task-type or construction-frame controls, or the separate 365-prompt extension.}
\label{fig:kink}
\end{figure}

The 95\% pairs-bootstrap interval, $[10.5,14.25]$, spans the stretch between the trough and the peak in Figure~\ref{fig:kink}; the bootstrap distribution is bimodal, with most draws at 14.0 and a smaller mode near 10.5 (Appendix~\ref{app:robustness}). The estimate describes this constructed benchmark, not a point at which harder prompts become easier.

\subsection{Task-Type and Construction-Frame Sensitivity}\label{sec:frame}

We label all 5{,}000 prompts with a nine-category taxonomy fixed before labeling; one out-of-panel judge labels every prompt, and three more label a shared 500-prompt subset (Krippendorff $\alpha=0.787$). With task-type fixed effects the breakpoint stays at 14.0 and the search still finds a break there (sup-Wald 78.6; 0 of 300 wild-bootstrap draws exceed it), but the task-adjusted jump across it is only $+0.4$ points ($p=0.66$), against $+6.8$ unadjusted (Table~\ref{tab:task-type}). The reason is plain: 91\% of prompts above 14.0 are data-structures/algorithms tasks, against 22\% below. Letting task effects also differ by side moves the break to 11.5, where the raw gap is 2.1 points. A piecewise term at 11.5 raises predictive $R^2$ from 0.093 for task type alone to 0.160, which is incremental fit, not causal independence. Four of five large task categories still reject the full IV restrictions, so task type does not explain away the overidentification failure.

\begin{table}[h]
\centering
\caption{Mean-pooled breakpoint under composition controls. The raw jump is the difference in mean pass rate across the selected break; the adjusted jump is the coefficient on $\mathbf{1}[C_i>\hat{\gamma}]$ in a regression on that indicator and the row's controls. Both are in percentage points. Every row's search rejects the no-break null with 0 of 300 wild-bootstrap exceedances (the unadjusted row also has 0 of 2{,}000; Section~\ref{sec:pooled}).}
\label{tab:task-type}
\begin{tabular}{@{}lrrrr@{}}
\toprule
Specification & $\hat{\gamma}$ & sup-Wald & Raw jump & Adjusted jump \\
\midrule
Unadjusted & 14.00 & 151.2 & $+6.8$ & $+6.8$ \\
Task-type fixed effects & 14.00 & 78.6 & $+6.8$ & $+0.4$ \\
Construction-frame fixed effect & 14.25 & 88.4 & $+6.8$ & $+1.8$ \\
Both & 14.25 & 51.1 & $+6.8$ & $-2.0$ \\
Task effects differing by side & 11.50 & 26.9 & $+2.1$ & n/a \\
\bottomrule
\end{tabular}
\end{table}

The construction frames tell a similar story. The 2{,}246 earlier prompts average composite 7.92 and pass rate 0.795; the 2{,}754 later ones, 11.42 and 0.897. A frame fixed effect leaves the break at 14.25 but cuts the adjusted jump to 1.8 points, and with both controls it turns slightly negative. Within frames, the earlier frame's best split, 9.50, is not significant (sup-Wald 4.02, bootstrap $p=0.14$) and a straight line fits better by BIC; the later frame breaks at 14.25 (sup-Wald 93.70, $p_{\mathrm{MC}}\leq1/301$) but rises only 1.8 points (0.890 to 0.908). Later prompts are 41.0\% of those at or below the pooled break and 95.4\% of those above it. So the dip and rebound are real features of this benchmark, but the big jump mostly reflects which source dominates the high region. The frame also marks separate generation runs and the test-quality difference noted in Section~\ref{sec:data}, so it is not purely a difference in prompts.

\subsection{Model and Pooling Heterogeneity}

Model-specific thresholds range from 10.0 to 14.25 (median 10.75). Nineteen models have higher mean pass above their own breakpoint and two lower: Kimi K2.5 (0.817 to 0.801 at 14.0) and Gemini 3.1 Pro Preview (0.872 to 0.793 at 10.75). Gemini's decline comes from 156 generations for which the API returned nothing (time-outs and empty responses in one batch, 144 of them at composite $>14$). Treated as missing rather than as failures, they turn Gemini's contrast upward (0.859 to 0.876 at 14.25), leave a small later-frame decline (0.890 to 0.869), and leave the pooled result unchanged (14.0; 0.834 to 0.907; Appendix~\ref{app:no-response}). How many models decline also depends on the outcome: five under the submission audit and one on raw test outcomes (Appendix~\ref{app:outcome-audit}).

Kimi K2.5 is the clearest decline. Its within-frame search is post hoc: we ran those searches only for models whose pass rate fell across their break (Section~\ref{sec:breakpoint}). With a frame fixed effect its search selects 14.25 (sup-Wald 45.5; 0 of 300 bootstrap draws exceed it), and so does a separate search within the later frame (sup-Wald 41.62), where its pass rate falls from 0.848 to 0.808 (95\% pairs-bootstrap interval $[14.00,14.75]$, whose upper end is the largest split the 500-prompt minimum allows, so it is truncated above). Piecewise BIC is 2{,}284 against 2{,}350 for a line. The earlier Kimi frame shows no significant break ($p_{\mathrm{MC}}=0.226$). Both 14.25 selections hold on raw test outcomes and under the submission audit.

The same prompts, scores, and tests are reused, so these are comparisons, not independent replications. The pooled threshold stays at 14.0 in all 21 leave-one-model-out fits; median pooling selects 10.5 and narrows the gap to 2.9 points (0.879 to 0.908). Details are in Appendix~\ref{app:per-model}; a library-mention diagnostic on a separate sample is in Appendix~\ref{app:library}.

\subsection{Audit-Clean High-Complexity Extension}

\begin{figure}[h]
\centering
\includegraphics[width=\linewidth]{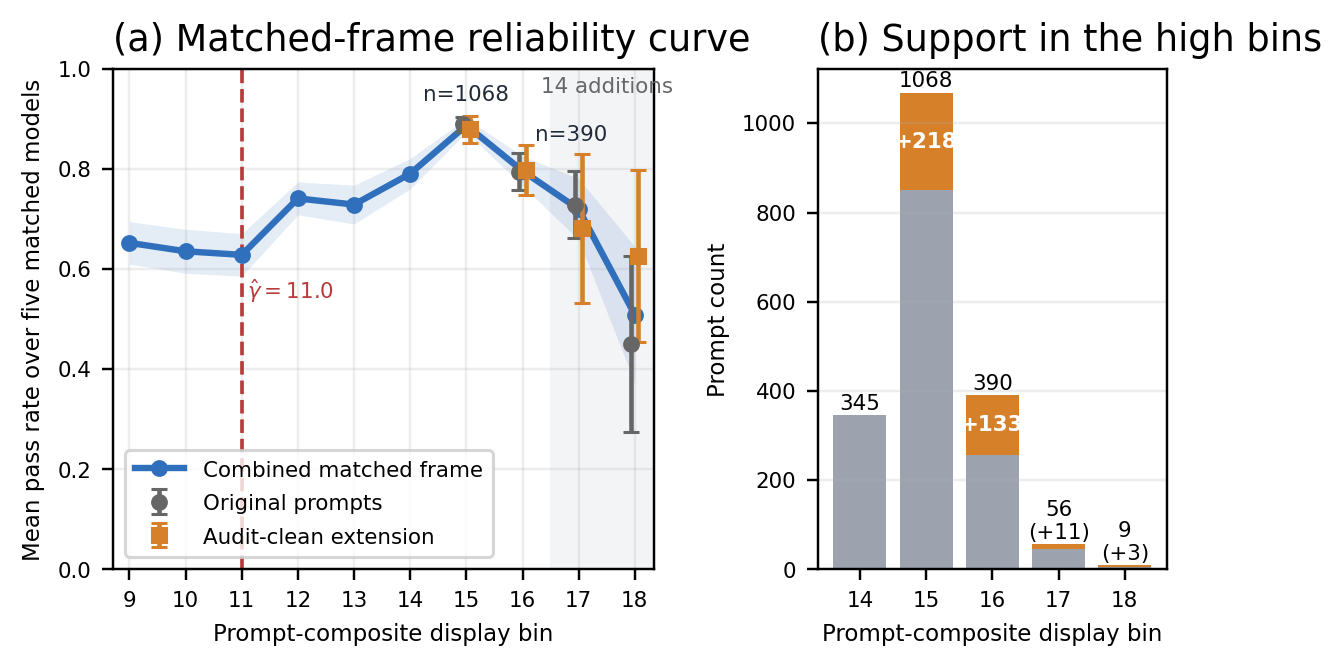}
\caption{Audit-clean extension in a matched five-model frame, on raw harness outcomes. (a) Pass rate is first averaged over the five models shared by the original benchmark and extension, then over prompts within display bins, where bin $b$ contains $C_i\in[b-0.5,b+0.5)$. Error bars and the shaded band are prompt-level normal-approximation intervals using $\pm1.96$ standard errors. The dashed line is the matched-panel threshold estimated on the unbinned composite. (b) The 365 additions materially raise support at bins 15 and 16. Only 14 additions lie above 16, so bins 17 and 18 remain descriptive.}
\label{fig:tail-extension}
\end{figure}

We separately selected high-composite candidates on the prompt side only, audited their contracts, and kept only prompts whose reference solution passes every test: 365 new prompts, 218/133/11/3 at display bins 15/16/17/18. Two caveats apply. Its generated code was not saved, so it could not be audited and we compare raw test outcomes on both sides. And it was generated under a different protocol, at temperature 0 with a system prompt that names the functions to define, which by itself can raise raw pass rates by removing naming failures. To hold the model mix fixed, Figure~\ref{fig:tail-extension} uses the five models in both runs (DeepSeek V3.2, GPT-OSS-120B, Kimi K2.5, Llama 3.3-70B, and Mistral Large-3). At bin 15 the extension and original means are 0.880 and 0.890; at bin 16, 0.799 and 0.795 (differences $-0.009$ and $+0.004$; Welch $p=0.554$ and $0.906$). Combined support rises to 1{,}068 and 390 prompts, and the matched-panel threshold on the unbinned composite is 11.0 (0.745 at or below, 0.810 above). That contrast pools both frames, so, as above, its upward gap reflects composition, not a rebound caused by complexity.

A separate fixed-version three-model check gives similar bin-15 and bin-16 rates (Appendix~\ref{app:tail}); with only 14 retained prompts above bin 16, the shape of the extreme tail remains open.

\subsection{Repeated-Sampling Robustness}

On a 359-prompt subset for four models (DeepSeek V3.2, GPT-OSS-120B, Kimi K2.5, and Llama 3.3-70B) with five draws per prompt-model cell at temperature 0.8 (7{,}180 generations on raw test outcomes; these were not part of the audit), the mean within-cell standard deviation is 0.064 and the median is zero. Requiring at least 59 prompts per side (a sixth of the subset), the first draw and the five-draw mean both select $\hat{\gamma}=14.25$; unlike the pooled 21-model contrast, pass rate here is lower above the break (first draw 0.788 to 0.755; Table~\ref{tab:passk}). Their correlation, 0.960, is a part-whole comparison, since the first draw is in the mean. This supports a stable threshold location for the sampled subset, not universal single-draw reliability.

% =====================================================================
\section{Discussion}
% =====================================================================

\textbf{The IV diagnostics.} The six-dimension IV specification does not support causal identification. Sargan $J=552.16$ on five degrees of freedom and the robust Wooldridge statistic (408.52) reject at machine precision, and not because of sample size: mean $J$ is already 33.4 at $n=250$ (null expectation about five), where every random subsample rejects. Single-instrument estimates disagree in sign (data structures $+0.0022$; state $-0.0040$; composition $-0.0045$), rotating the dimensions does not fix that, and the few passing subsets were picked after seeing the data (Appendix~\ref{app:econ-diagnostics}). What survives is the composite as an index measured from the prompt alone: output failure cannot reach it, and it can be checked directly against reference CC, human ratings, paraphrases, and re-expressions. None of those checks makes it an instrument.

\textbf{What the breakpoint means.} The pooled breakpoint compactly describes a nonmonotone curve in a selection-stratified benchmark; it is not a universal difficulty limit. Task-type fixed effects keep its location but remove nearly all of its jump, neither frame reproduces the jump, median pooling moves it to 10.5, the outcome definition moves it between 11.25 and 14.0, and model directions differ. Leave-one-model-out stability shows that no single model drives it, not that it is independent of pooling, composition, or the outcome definition. Kimi K2.5 is the clearest downward, benchmark-conditional case: its later-frame break and decline hold under all three outcome definitions. A model's $\hat{\gamma}$ is therefore a candidate reliability breakpoint for that model under this frame, index, generation setting, and test protocol. The extension adds evidence only where it adds support: on raw outcomes it matches bins 15 and 16, and the tail beyond them is open.

\textbf{Why a prompt-side index, and implications.} The central design choice is what the index is allowed to see: the input only, never the output. A lexical proxy shares that advantage and may be preferable when transparency or cost matters most; the rubric trades cost for interpretable dimensions and measurable rater disagreement, and the two-grader calibration argues for reporting that uncertainty rather than trusting any single rater. The idea may carry beyond Python, but our evidence covers only score transfer to Java and C++. If you bin results by output complexity, say how failed and missing outputs are counted. Prompt-side scores can help, as long as you also show how prompts were sampled, how reliable the raters are, how many prompts each bin holds, and pooled and task-adjusted results. They are diagnostic summaries, not deployment rules, until validated prospectively.

% =====================================================================
\section{Limitations}
\label{sec:limitations}
% =====================================================================

\textbf{Constructed sampling frame.} The benchmark is balanced on a preliminary single-rater score, not the final index, and inherits a 200{,}000-record prefix scan and deliberate high-reference-complexity supplementation. The two frames differ sharply in score, pass rate, test quality, and generation run, and their changing mixture explains much of the pooled rebound. The benchmark does not estimate performance under the natural OpenCodeInstruct distribution, and every estimate is conditional on the model or pooling rule, frame, index, generation settings, and test protocol.

\textbf{Composition and pooling.} Task-type controls remove most of the pooled jump, the within-frame changes are small or not significant, and median pooling moves the break; read the pooled 14.0 only with those sensitivities.

\textbf{Outcome definition.} The outcome depends on one LLM auditor, adopted by author decision after no candidate met our pre-declared rule: it missed the clean-accuracy threshold by three cases during selection, then met every threshold on a fresh set, a check that no longer gated adoption. A second auditor agrees with it at $\kappa=0.631$, and its known-answer accuracy is measured against labels that are themselves noisy. The extension and repeated-sampling checks use raw test outcomes, and the 157 generations with no API response count as failures (Appendices~\ref{app:outcome-audit} and~\ref{app:no-response}).

\textbf{Scorer dependence and human disagreement.} Common-mode LLM bias can carry into the index, and the judges are out-of-panel but not out-of-family (GPT-5.5 and Llama 4 share developers with panel models); the ensemble makes disagreement visible but cannot remove it. Human-LLM agreement is only moderate on the disagreement-enriched sample and varies by rater.

\textbf{No causal IV claim.} The overidentification restrictions reject even in small subsamples and the candidate instruments imply conflicting coefficients, so the 2SLS fits cannot identify a causal effect.

\textbf{Scope.} AuroraGPT-IT-v4 was excluded post hoc without a prespecified rule, so a like-for-like 22-model sensitivity is unavailable. The extension adds 351 prompts at bins 15 and 16 but only 14 above 16, leaving that endpoint underpowered. Generation and execution are Python-only; the Java and C++ results test score transfer only. The task taxonomy, frame analyses, extension, human, paraphrase, language, and repeated-sampling checks were added post hoc and not preregistered, and we do not track prompts or model families over time.

% =====================================================================
\section{Conclusion}
% =====================================================================

Generated-output cyclomatic complexity is a poor stand-in for prompt structure when failures produce short or unscorable programs. A prompt-side index avoids that problem and supports model-specific breakpoint estimation; the clearest downward example, found in post hoc follow-up, is Kimi K2.5 near 14.25, which holds under a frame control, within the later frame, and across outcome definitions. The pooled result is less stable: task type removes most of its jump, the prompt source explains the rebound, and pooling and the outcome definition move it. The extension matches bins 15 and 16 on raw outcomes, human calibration is moderate, paraphrase, language, and repeated-sampling checks preserve ordering and location, and the candidate instruments fail their joint restrictions, so the contribution is measurement and observational analysis, not causal identification. The lesson for benchmark designers is modest: add prompt-only measures to output-derived ones, publish the rubric and rater uncertainty, show support counts, and report how composition, pooling, and the outcome definition change the curve.

\vspace{4pt}
\noindent\textbf{Data and code availability.} Code, the rubric, and figure source tables are at \codeurl{}. The benchmark release, with Croissant metadata, is at \dataurl{} (CC BY 4.0): prompts, judge scores, all 105{,}000 generations with three outcome definitions, audit verdicts, and Lizard complexity, the auditor's known-answer sets, and every supporting set (Appendix~\ref{app:artifact}).

% =====================================================================
% Acknowledgments and funding disclosure: required in the camera-ready version;
% like references, this section does not count toward the content-page limit.
% Funding and competing-interest statements confirmed by the first author, 2026-09-30.
\begin{ack}
We thank the Connected Systems Institute at the University of Wisconsin -- Milwaukee for providing the computing resources for this work. Rubric scoring, unit-test execution, and the Azure-hosted model inference ran on a Microsoft Azure subscription funded by the Connected Systems Institute. All other model and auditor API usage was paid for personally by the first author. % No grants supported this work.

\textbf{Competing interests.} The authors declare no competing interests.
\end{ack}

\appendix

\section{Workflow and Robustness Designs}
\label{app:robustness-designs}

\begin{figure}[h]
\centering
\includegraphics[width=\linewidth]{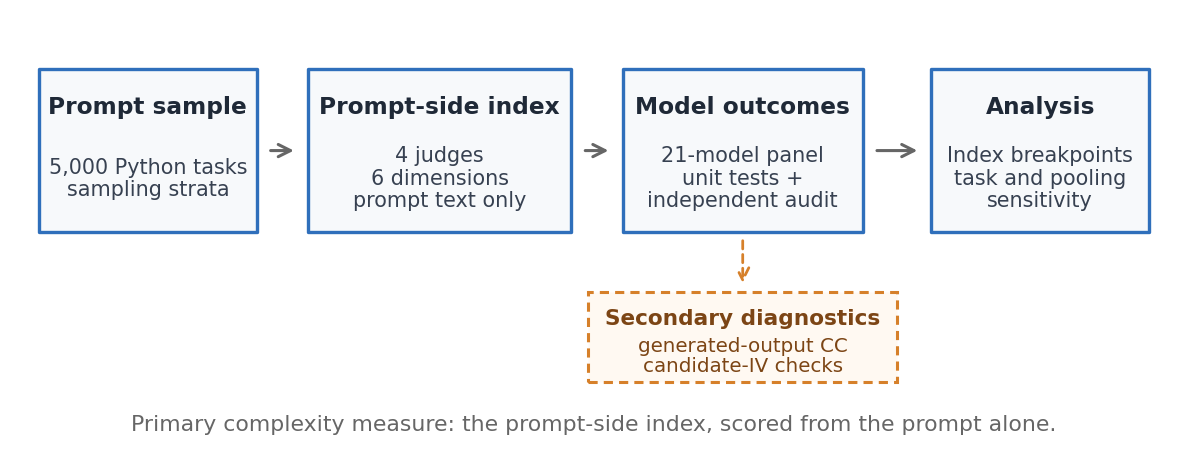}
\caption{Measurement and analysis workflow. The prompt-side composite is computed from the prompt text alone, without access to generations or test results, and is primary in the breakpoint analysis. Unit-test verdicts are reviewed by an independent auditor (Appendix~\ref{app:outcome-audit}). Generated-output complexity and 2SLS fits are retained as diagnostic comparisons.}
\label{fig:pipeline}
\end{figure}

Figure~\ref{fig:pipeline} summarizes the measurement and analysis workflow.

All of the following stress tests were added after the initial analysis and are post hoc robustness checks, not preregistered confirmation. Nine task categories are assigned to all prompts by one out-of-panel judge; three additional judges label a shared 500-prompt subset. We examine the two recorded prompt-source frames, compare first-draw and five-draw outcomes on 359 prompts and four models, and re-express 117 prompts in each of Java and C++ for rescoring. For the paraphrase check, we sample 150 prompts, 112 drawn from rounded composite bins 10--18 and the remaining 38 drawn outside that region. A non-judge DeepSeek-V3.2 model is instructed to preserve the exact required inputs, outputs, behavior, constraints, and edge cases while removing only verbose framing; the same four judges then rescore the rewrites. Both authors grade a disagreement-enriched calibration sample with the LLM scores hidden. A prompt-side-selected extension retains 365 new prompts only after contract audit and successful execution of every reference solution. Finally, we rerun the main fits treating the 157 generations for which the model API returned nothing as missing rather than as failures (Appendix~\ref{app:no-response}).

\section{Rubric Alignment Details}
\label{app:alignment}

Table~\ref{tab:rubric-ref-cc} reports the composite distribution by reference-solution $\kappa$ bin, and Table~\ref{tab:low-end-separation} the low-end comparison for two of the eight keyword counts computed from the same prompt text (all eight are in the release): a total count of structural cue words and a loop-specific cue count. Like the rubric, the keyword features use only the prompt text, but they detect only surface words. The rubric's $\kappa=1$ versus $\kappa=2$ separation is driven primarily by iteration ($+0.24$), state ($+0.19$), and branching ($+0.09$).

\begin{table}[h]
\centering
\caption{Rubric composite by reference-solution cyclomatic complexity bin for all 5{,}000 prompts. The two measures have a strong positive association overall (Pearson $r=0.80$, Spearman $\rho=0.81$), although the bin means are not monotone at every adjacent value.}
\label{tab:rubric-ref-cc}
{\small
\setlength{\tabcolsep}{4pt}
\begin{tabular}{@{}lccccccccccccc@{}}
\toprule
Reference $\kappa$
& 1 & 2 & 3 & 4 & 5 & 6 & 7 & 8 & 9 & 10 & 11 to 15 & 16 to 20 & 21 to 50 \\
\midrule
$N$
& 357 & 338 & 324 & 300 & 301 & 153 & 136 & 184 & 89 & 75 & 308 & 876 & 1559 \\
Mean
& 4.64 & 5.25 & 6.23 & 7.25 & 6.61 & 7.15 & 8.08 & 7.75 & 8.79 & 7.44 & 9.52 & 10.71 & 14.34 \\
SD
& 2.39 & 2.07 & 2.19 & 2.43 & 2.78 & 2.86 & 2.94 & 3.12 & 2.88 & 3.41 & 3.02 & 3.45 & 1.64 \\
\bottomrule
\end{tabular}
}
\end{table}

\begin{table}[h]
\centering
\caption{The prompt-side rubric discriminates $\kappa=1$ from $\kappa=2$ from the prompt alone. Keyword features capture some surface structure but are less directly tied to intended code structure.}
\label{tab:low-end-separation}
\begin{tabular}{@{}lrrrr@{}}
\toprule
Feature & $\kappa=1$ & $\kappa=2$ & $\Delta$ & $p$-value \\
\midrule
Rubric composite                  & 4.64 & 5.25 & $+0.61$ & $3.1\times10^{-4}$ \\
Keyword structural count          & 2.76 & 3.34 & $+0.58$ & $2.4\times10^{-3}$ \\
Keyword loop count                & 0.07 & 0.04 & $-0.02$ & n.s. \\
\bottomrule
\end{tabular}
\end{table}

\section{Additional Econometric Diagnostics}
\label{app:econ-diagnostics}

\subsection{First-stage relevance}

The first-stage regression (\ref{eq:stage1}) on the combined dataset yields a heteroscedasticity-robust Wald statistic of $\chi^2(6)=13{,}611.3$ and an HC1-robust joint $F(6,4993)=2{,}265.4$ (classical $F=1{,}779.4$). Partial $R^2$ on generated-output complexity is $0.681$. Per-model robust Wald statistics range from $\chi^2(6)=6{,}053.7$ (GPT-5-mini) to 14{,}693.2 (Devstral Small 2505). The dimensions are strongly relevant to output CC, but relevance does not establish exclusion validity.

\begin{table}[h]
\centering
\caption{First-stage OLS coefficients for candidate rubric dimensions predicting generated-output complexity. Heteroscedasticity-consistent (HC1) standard errors are in parentheses. $N=5{,}000$.}
\label{tab:stage1}
\begin{tabular}{@{}llc@{}}
\toprule
Dimension & $\hat{\pi}$ & SE \\
\midrule
Branching        & $1.090^{***}$  & $(0.238)$ \\
Iteration        & $0.634^{***}$  & $(0.172)$ \\
State            & $2.091^{***}$  & $(0.292)$ \\
Data structures  & $2.851^{***}$  & $(0.138)$ \\
Edge cases       & $6.024^{***}$  & $(0.279)$ \\
Composition      & $-0.766^{**}$  & $(0.291)$ \\
Constant         & $-4.924^{***}$ & $(0.311)$ \\
\midrule
Partial $R^2$                & \multicolumn{2}{c}{0.681} \\
HC1-robust joint $F$         & \multicolumn{2}{c}{2{,}265.4} \\
Robust Wald $\chi^2(6)$      & \multicolumn{2}{c}{13{,}611.3} \\
$N$                          & \multicolumn{2}{c}{5{,}000} \\
\bottomrule
\multicolumn{3}{l}{\footnotesize $^{**}p<0.01$, $^{***}p<0.001$}
\end{tabular}
\end{table}

All six dimensions are individually significant first-stage predictors. \emph{Edge cases} and \emph{data structures} have the largest conditional associations; \emph{composition} is negative after conditioning on the other five dimensions, illustrating substantial shared variance.

\subsection{Overidentification diagnostics}

The Sargan statistic is $552.16$ on $\chi^2(5)$ and the robust Wooldridge statistic is $408.52$; both reject at machine precision. The same pattern appears in every model-specific fit. The restrictions fail well before the full sample size. Table~\ref{tab:overid-subsamples} summarizes 200 random draws at each $n$.

\begin{table}[h]
\centering
\caption{Overidentification under random subsampling. Under the null, $\mathbb{E}[J]\approx5$. Values are means over 200 draws at each subsample size.}
\label{tab:overid-subsamples}
\begin{tabular}{@{}rrrr@{}}
\toprule
$n$ & Mean $J$ & Mean $J/n$ & Share rejecting at 0.05 \\
\midrule
250   & 33.35  & 0.1334 & 1.000 \\
500   & 61.03  & 0.1221 & 1.000 \\
1{,}000 & 114.57 & 0.1146 & 1.000 \\
2{,}000 & 224.58 & 0.1123 & 1.000 \\
3{,}000 & 332.78 & 0.1109 & 1.000 \\
4{,}000 & 443.53 & 0.1109 & 1.000 \\
5{,}000 & 552.16 & 0.1104 & 1.000 \\
\bottomrule
\end{tabular}
\end{table}

The just-identified coefficients disagree in sign: branching $-0.00174$ ($p=9.9\times10^{-7}$), iteration $-0.00092$ ($p=0.063$), state $-0.00402$ ($p=1.9\times10^{-18}$), data structures $+0.00220$ ($p=5.4\times10^{-9}$), edge cases $-0.00170$ ($p=7.1\times10^{-7}$), and composition $-0.00452$ ($p<10^{-15}$). The sign disagreement is consistent with the failed joint restrictions.

Principal-component rotation preserves the same disagreement. PC1 explains 78.5\% of the dimension variance and, as the sole instrument in the prompt-level pass-rate-on-mean-output-CC diagnostic, gives $\hat{\beta}=-0.00110$ (SE $=0.00032$, $p=0.0007$) with robust first-stage Wald $\chi^2(1)=8{,}489$. Because that fit is just-identified, it has no overidentification test and is a robustness specification rather than a repair. Adding PC2 raises cumulative explained variance to 87.7\% and already rejects ($J=66.23$, $p=4.0\times10^{-16}$); every larger leading-PC specification also rejects, and the full six-PC fit reproduces $J=552.16$ exactly. Under the reported component orientation, PC1 loads broadly across all six dimensions, while PC2 contrasts branching ($+0.54$) and edge cases ($+0.48$) against data structures ($-0.54$) and iteration ($-0.41$).

The post hoc search over 56 subsets of two to five dimensions finds four Sargan-nonrejecting specifications: branching plus edge cases ($p=0.821$), iteration plus edge cases ($p=0.052$), state plus composition ($p=0.110$), and branching plus iteration plus edge cases ($p=0.097$). Their first-stage statistics range from 4{,}652 to 7{,}768. These subsets were selected after examining the same data and are multiplicity-sensitive diagnostics, not validation of the original instrument set.

\subsection{Output-CC regression diagnostics}

\begin{table}[h]
\centering
\caption{Descriptive OLS and candidate-IV 2SLS fits of prompt-level mean pass rate, averaged over all 21 generations, on prompt-level mean generated-output CC, averaged over the generations with computable CC. Of 5{,}000 prompts, 4{,}294 have all 21 model values and 706 use fewer. OLS uses HC1 standard errors and 2SLS uses heteroscedasticity-robust standard errors. Because the full instrument set fails overidentification, the 2SLS coefficient is not causal.}
\label{tab:ols-vs-2sls}
\begin{tabular}{@{}lcc@{}}
\toprule
Regressor: generated-output CC & OLS & Candidate-IV 2SLS \\
\midrule
$\hat{\beta}$ & $+0.00036$ & $-0.00070^{*}$ \\
              & $(0.00029)$ & $(0.00032)$ \\
\midrule
Robust first-stage Wald $\chi^2(6)$ & n/a & 13{,}611.3 \\
Partial $R^2$ & n/a & 0.681 \\
$N$ & \multicolumn{2}{c}{5{,}000} \\
\bottomrule
\multicolumn{3}{l}{\footnotesize $^{*}p<0.05$}
\end{tabular}
\end{table}

The direct output-CC slope is near zero and not significant ($p=0.21$), while the invalid candidate-IV fit is small and negative ($p=0.025$). The failed exclusion restrictions prevent a causal reading of their difference. Separately, a fractional-probit fit of pass rate on the prompt composite gives a marginal effect at the mean of $-0.00304$ (coefficient $p=0.009$); linear OLS on the prompt composite gives $-0.00290$ ($R^2=0.0026$, $p=2.0\times10^{-5}$). These small global prompt-index slopes are compatible with the visibly nonmonotone conditional curve.

\subsection{Keyword-feature baseline}

Table~\ref{tab:keyword-baseline} compares the keyword-feature baseline and the prompt-side rubric on the main measurement weaknesses. The keyword baseline is an internal lexical proxy computed from prompt text, not a prior published result.

\begin{table}[h]
\centering
\caption{Keyword-feature baseline and prompt-side rubric. Both use only the prompt text; the rubric adds explicit dimensions and rater uncertainty, while the lexical proxy is cheaper and deterministic.}
\label{tab:keyword-baseline}
\begin{tabular}{@{}lll@{}}
\toprule
Issue & Keyword baseline & Prompt rubric \\
\midrule
Output-side contamination & Avoided & \textbf{Avoided} \\
Input used                & Prompt text only & Prompt text only \\
$\kappa=1$ vs $\kappa=2$  & $\Delta=+0.58$ & $\Delta=+0.61$ \\
Number of features        & 8 keyword counts & 6 dimensions \\
Scoring rule              & Deterministic counts & Four-rater mean \\
Scorer disagreement       & n/a     & observable \\
\bottomrule
\end{tabular}
\end{table}

\clearpage
\section{Per-model Details}
\label{app:per-model}

\begin{figure}[h]
\centering
\includegraphics[width=\linewidth]{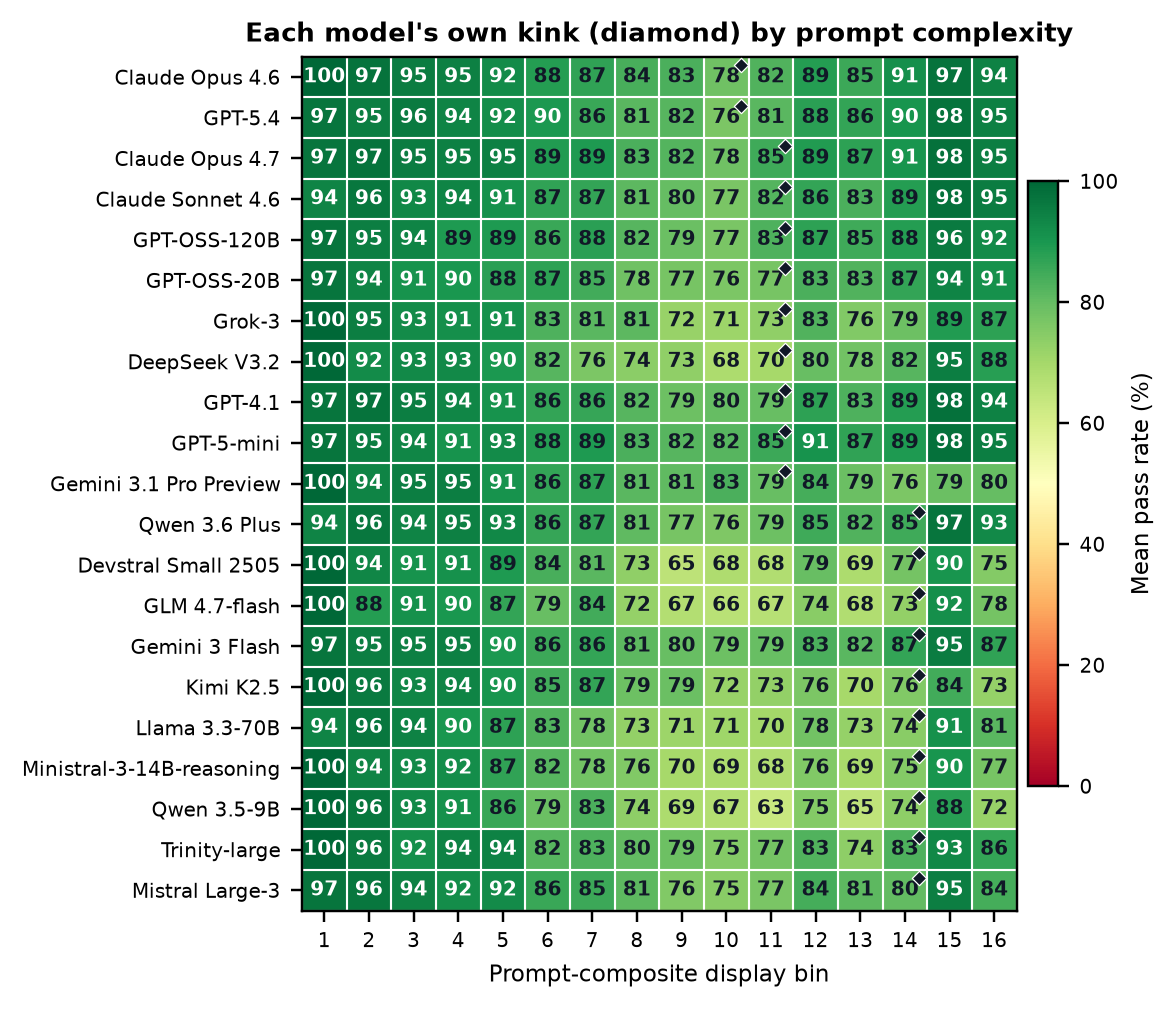}
\caption{Model-specific pass rate by prompt-side composite display bin in the original benchmark. Bin $b$ contains $C_i\in[b-0.5,b+0.5)$, so exact half-point boundaries enter the higher bin. Rows are 21 evaluated models and columns are bins 1 to 16. The diamond marks each model's fitted breakpoint in the corresponding display bin. Sparse original bins 0 and 17 to 19 contain 3, 45, 6, and 5 prompts and are omitted.}
\label{fig:heatmap}
\end{figure}

\begin{table}[h]
\centering
\caption{Unadjusted original-benchmark model-specific breakpoint estimates. $\Delta$ is Pass$_{C>\hat{\gamma}}-$Pass$_{C\leq\hat{\gamma}}$ in percentage points. Boot and placebo columns report exceedances out of 500 resamples (placebo: shuffled index); zero exceedances imply finite Monte Carlo $p_{\mathrm{MC}}\leq1/501\approx0.002$, not $p=0$, and three exceedances give $p_{\mathrm{MC}}=4/501\approx0.008$. The combined row uses 2{,}000 resamples. These before/after summaries are not construction-frame-stable failure onsets. $^\ddagger$Gemini 3.1 Pro Preview's decline comes from 156 generations with no API response; treated as missing, its row is 14.25, 85.9\%, 87.6\% (Appendix~\ref{app:no-response}).}
\label{tab:per-model}
\small
\begin{tabular}{@{}lrrrrrr@{}}
\toprule
Model & $\hat{\gamma}$ & Pass $C\leq\hat{\gamma}$ & Pass $C>\hat{\gamma}$ & $\Delta$ & Boot exc. & Plcb exc. \\
\midrule
Claude Opus 4.6          & 10.00 & 88.6\% & 91.5\% & +2.8 & 0/500 & 0/500 \\
Claude Opus 4.7          & 10.50 & 88.8\% & 92.6\% & +3.8 & 0/500 & 0/500 \\
Claude Sonnet 4.6        & 10.50 & 86.5\% & 91.3\% & +4.7 & 0/500 & 0/500 \\
Trinity-large            & 14.00 & 83.4\% & 90.3\% & +7.0 & 0/500 & 0/500 \\
DeepSeek V3.2            & 10.75 & 81.0\% & 85.7\% & +4.7 & 0/500 & 0/500 \\
GPT-OSS-120B             & 10.50 & 85.9\% & 90.5\% & +4.6 & 0/500 & 0/500 \\
Grok-3                   & 10.50 & 83.2\% & 83.3\% & +0.1 & 0/500 & 0/500 \\
Kimi K2.5                & 14.00 & 81.7\% & 80.1\% & $-1.6$ & 0/500 & 0/500 \\
Llama 3.3-70B            & 14.00 & 79.0\% & 86.0\% & +7.0 & 0/500 & 0/500 \\
Mistral Large-3          & 14.25 & 84.1\% & 91.1\% & +6.9 & 0/500 & 0/500 \\
GLM 4.7-flash            & 14.00 & 77.0\% & 85.9\% & +8.9 & 0/500 & 0/500 \\
Gemini 3 Flash           & 14.00 & 85.4\% & 92.0\% & +6.6 & 0/500 & 0/500 \\
Gemini 3.1 Pro Preview$^\ddagger$ & 10.75 & 87.2\% & 79.3\% & $-7.9$ & 3/500 & 0/500 \\
GPT-4.1                  & 10.75 & 86.7\% & 91.2\% & +4.6 & 0/500 & 0/500 \\
GPT-5-mini               & 10.75 & 87.9\% & 92.8\% & +4.9 & 0/500 & 0/500 \\
GPT-OSS-20B              & 10.50 & 84.3\% & 88.1\% & +3.7 & 0/500 & 0/500 \\
Ministral-3-14B-reasoning & 14.00 & 78.1\% & 84.6\% & +6.4 & 0/500 & 0/500 \\
Devstral Small 2505      & 14.00 & 78.7\% & 84.5\% & +5.8 & 0/500 & 0/500 \\
GPT-5.4                  & 10.00 & 87.8\% & 91.2\% & +3.4 & 0/500 & 0/500 \\
Qwen 3.5-9B              & 14.00 & 77.3\% & 81.7\% & +4.4 & 0/500 & 0/500 \\
Qwen 3.6 Plus            & 13.75 & 85.2\% & 94.5\% & +9.3 & 0/500 & 0/500 \\
\midrule
Combined (21 models)     & 14.00 & 83.4\% & 90.1\% & +6.8 & 0/2{,}000 & 0/2{,}000 \\
\bottomrule
\end{tabular}
\end{table}

\clearpage
\section{Reverse-threshold and Robustness Details}
\label{app:robustness}

\subsection{Bootstrap and placebo settings}

The mean-pooled analysis uses 2{,}000 wild-bootstrap draws, 1{,}000 pairs-bootstrap draws for the interval, and 2{,}000 shuffled-index placebo draws. Unadjusted model-specific fits use 500 wild-bootstrap and 500 placebo draws. Every controlled search, additive or differing by side, pooled or model-specific, uses a 300-draw wild bootstrap under its own no-break model (seed 42), and the within-frame threshold intervals use 300 pairs-bootstrap resamples. Pairs-bootstrap intervals are percentile intervals and are descriptive: for a threshold location they are not guaranteed to have nominal coverage, and the headline bootstrap distribution is bimodal (most draws at 14.0, a smaller mode near 10.5). The repeated-sampling search on 359 prompts requires at least 59 prompts on each side (one sixth of the subset) instead of 500.

\subsection{Reverse-threshold complete cases}

\begin{figure}[h]
\centering
\includegraphics[width=\linewidth]{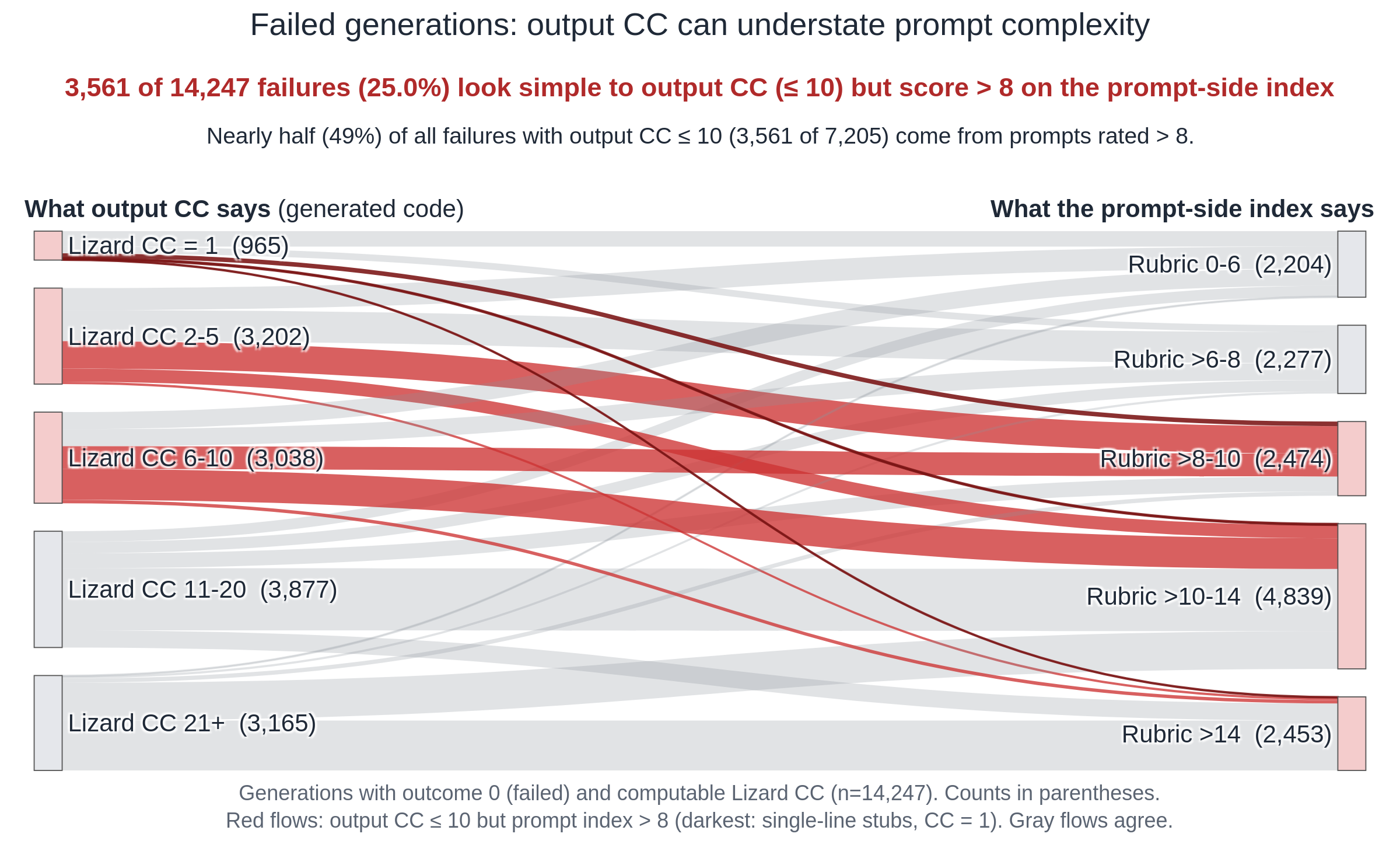}
\caption{Reverse Threshold Problem among 14{,}247 failed generations (outcome 0) with computable Lizard CC. Generated-output complexity is on the left and the prompt composite is on the right. Highlighted flows have output CC $\leq10$ and prompt composite $>8$.}
\label{fig:sankey}
\end{figure}

Figure~\ref{fig:sankey} expands the 2-by-2 diagnostic in Figure~\ref{fig:output-cc}. Failed complete cases (outcome 0) flow from Lizard complexity on generated code to the prompt composite. Highlighted flows are prompts with composite $>8$ whose failed outputs have CC $\leq10$. Under the audited outcome, ``failed'' is the auditor's verdict: of these 14{,}247 generations, 4{,}215 passed no unit test, 9{,}222 passed some, and 810 passed all of them but were judged incorrect.

In the combined analysis, none of 2{,}000 wild-bootstrap or 2{,}000 shuffled-index placebo statistics reaches the observed $W^*=151.15$. With the standard plus-one correction, each gives $p_{\mathrm{MC}}\leq1/2001<0.001$. For each unadjusted model-specific fit, the observed statistic exceeds all 500 placebo draws and all 500 bootstrap draws ($p_{\mathrm{MC}}\leq1/501\approx0.002$), except that three bootstrap draws exceed it for Gemini 3.1 Pro Preview ($p_{\mathrm{MC}}=4/501\approx0.008$).

\subsection{Task-type and control specifications}

The nine fixed categories contain 2{,}005 data-structures/algorithms prompts, 654 array/matrix/grid, 619 numeric/math, 617 string/text, 341 other, 276 stateful simulation, 215 parsing/serialization, 190 logic/validation, and 83 API/library orchestration prompts. On the shared 500-prompt reliability set, pairwise Cohen $\kappa$ ranges from 0.737 to 0.856 and Krippendorff $\alpha=0.787$.

The two control specifications answer different questions. With additive fixed effects the break model is $y_i=\alpha+\beta C_i+\mathbf{x}_i'\boldsymbol{\delta}+\mathbf{1}[C_i>\gamma](\alpha'+\beta'(C_i-\gamma))+u_i$, so only the composite line may break. The by-side specification (the submitted analysis's) fits separate regressions, controls included, on each side, so its test also asks whether task or frame effects differ across the break; that is why it moves the task-controlled break to 11.5 (where the task-adjusted jump is $-6.2$ points) and, under the submission-audit and raw-harness outcomes, the frame-controlled break to 8.50 (Table~\ref{tab:outcome-sensitivity}). By side, the frame-controlled search selects 14.25 (sup-Wald 59.0). Adjusted jumps at the unadjusted $\hat{\gamma}=14.0$ are $+6.8$ with no controls, $+0.4$ with task type ($p=0.66$), $+1.6$ with the frame ($p=0.03$), and $-2.8$ with both ($p=0.001$).

The piecewise composite at 11.5 has $R^2=0.04530$ alone. Task-type fixed effects have $R^2=0.09285$ alone, and their combination has $R^2=0.16010$. Thus the piecewise term adds 0.06725 in-sample predictive $R^2$ beyond the labels. Table~\ref{tab:within-type-j} shows that the IV failure is not localized to one task category.

\begin{table}[h]
\centering
\caption{Full six-dimension Sargan test within task categories having at least 300 prompts.}
\label{tab:within-type-j}
\begin{tabular}{@{}lrrrr@{}}
\toprule
Task category & $n$ & $J$ & $p$ & $J/n$ \\
\midrule
Data structures/algorithms & 2{,}005 & 208.81 & $<10^{-15}$ & 0.1041 \\
Array/matrix/grid          & 654     & 86.23  & $4.1\times10^{-17}$ & 0.1319 \\
Numeric/math               & 619     & 62.53  & $3.6\times10^{-12}$ & 0.1010 \\
String/text                & 617     & 39.70  & $1.7\times10^{-7}$  & 0.0643 \\
Other                      & 341     & 10.89  & 0.054 & 0.0319 \\
\bottomrule
\end{tabular}
\end{table}

\subsection{Construction-frame sensitivity}

All 5{,}000 prompts map to the recorded construction frame. The retained earlier frame contains 2{,}246 prompts and the later candidate frame contains 2{,}754. At the pooled $\hat{\gamma}=14.0$, later candidates account for 41.0\% of the lower regime and 95.4\% of the upper regime. Within the earlier frame, raw pass rate is 0.795 at or below versus 0.795 above that pooled threshold, with only 60 prompts above. Within the later frame, it is 0.890 versus 0.907. The overlap limitation makes reweighting unstable, so we report the controlled and within-frame fits rather than treating a standardized contrast as definitive.

\begin{table}[h]
\centering
\caption{Unbinned threshold searches within the two construction frames. Regime means are raw summaries at each frame's selected breakpoint.}
\label{tab:source-frame}
\begin{tabular}{@{}lrrrrrr@{}}
\toprule
Frame & $n$ & $\hat{\gamma}$ & sup-Wald & Bootstrap $p$ & Pass low & Pass high \\
\midrule
Retained earlier & 2{,}246 & 9.50 & 4.02 & 0.14 & 0.835 & 0.704 \\
Later candidate & 2{,}754 & 14.25 & 93.70 & $\leq1/301$ & 0.890 & 0.908 \\
\bottomrule
\end{tabular}
\end{table}

For the earlier frame, linear BIC is 777.1 versus 784.5 for the piecewise fit. For the later frame, piecewise BIC is $-1{,}626.4$ versus $-1{,}460.6$ for linear. Across both frames, a model with frame-specific linear slopes has $R^2=0.0856$ and BIC $-207.8$, compared with $R^2=0.0955$ and BIC $-245.7$ for the by-side frame-controlled breakpoint model; frame-specific slopes plus a common piecewise term at 14.25 fit best ($R^2=0.109$, BIC $-320.6$). Thus nonlinearity survives frame control, but the later frame's significant change (+1.8 points) is far smaller than the pooled gap, and the earlier frame's larger raw drop is not a significant break. The frame indicator also marks separate generation runs (weeks apart, sometimes with different routes or token limits) and test quality: for 1{,}259 earlier prompts the source reference solution fails at least one of its tests (491 fail all), while every later prompt's reference passes.

\subsection{Pooling and repeated sampling}

Every leave-one-model-out mean-pooled fit selects 14.0. Median pooling selects 10.5, with 0.879 at or below and 0.908 above. Table~\ref{tab:passk} gives the repeated-sampling subset: 359 prompts, four models, five draws, and 7{,}180 executed generations. Its code was saved, but these generations were not part of the independent audit, so outcomes are raw harness values; they were generated at temperature 0.8 with the extension's system prompt (Appendix~\ref{app:tail}). The within-cell standard deviation uses divisor five (population SD), and the threshold search requires at least 59 prompts per side.

\begin{table}[h]
\centering
\caption{Repeated-sampling threshold check. The first draw is part of the five-draw mean, so $r=0.960$ is a part-whole comparison.}
\label{tab:passk}
\begin{tabular}{@{}lrrrr@{}}
\toprule
Outcome & $\hat{\gamma}$ & sup-Wald & Pass $C\leq\hat{\gamma}$ & Pass $C>\hat{\gamma}$ \\
\midrule
First draw & 14.25 & 10.63 & 0.788 & 0.755 \\
Mean of five draws & 14.25 & 11.42 & 0.785 & 0.750 \\
\bottomrule
\end{tabular}
\end{table}

\subsection{Library-mention diagnostic}
\label{app:library}

A 459-prompt diagnostic at display bins 13 and 17 tags case-insensitive whole-word matches to a fixed list of library, framework, and standard-library names. An HC1 regression of prompt-level mean pass on that tag controls for source-pool membership, prompt characters and words, unit-test count and length, reference-solution CC, scorer disagreement, and the composite. The mention-tag coefficient is $-0.0501$ (robust SE $=0.0978$, $p=0.608$). This sample is not a subset of the analyzed benchmark: it combines 267 benchmark prompts with 192 prompts from a separate candidate pool that was mined on generated-output CC, and it uses submission-audit outcomes, so it cannot be re-estimated under the independent audit. Within the benchmark, too few prompts at these bins carry the tag to analyze. The diagnostic therefore offers no support for a library or framework explanation, but it is weak evidence either way.

\subsection{High-complexity extension}
\label{app:tail}

Table~\ref{tab:tail-detail} keeps the original and extension sources separate. Pass rate is first averaged over the five matched models within prompt, using raw harness outcomes for both sources. The two well-supported comparisons differ by about one percentage point or less and are not statistically distinguishable. Bins 17 and 18 remain too small for stable comparisons. The extension runs used temperature 0.0, at most 4{,}096 output tokens, and a system prompt that tells the model to ``define exactly the function(s) or class(es) the task names, with the specified signature''; the main runs did not include that instruction. Since naming mismatches are a common source of raw test failures, the protocol difference by itself could raise extension pass rates, so the close match at bins 15 and 16 is weaker evidence than a same-protocol replication.

\begin{table}[h]
\centering
\caption{Matched-five-model high-bin support and raw-harness pass rates. Parentheses contain prompt counts.}
\label{tab:tail-detail}
\begin{tabular}{@{}lrrrr@{}}
\toprule
Source & Bin 15 & Bin 16 & Bin 17 & Bin 18 \\
\midrule
Original & 0.890 (850) & 0.795 (257) & 0.729 (45) & 0.450 (6) \\
Audit-clean extension & 0.880 (218) & 0.799 (133) & 0.682 (11) & 0.627 (3) \\
Combined & 0.888 (1{,}068) & 0.797 (390) & 0.720 (56) & 0.509 (9) \\
\midrule
Welch $p$, original vs extension & 0.554 & 0.906 & n/a & n/a \\
\bottomrule
\end{tabular}
\end{table}

The separate fixed-version three-model check, also on raw harness outcomes, analyzes 515 prompts per model: 150 midrange anchors plus 365 audit-clean additions. Claude Opus 4.6, GPT-5.4, and Gemini 3.1 Pro Preview have bin-15 pass rates 0.941, 0.940, and 0.936, and bin-16 rates 0.871, 0.850, and 0.845. The 11 and three prompts in bins 17 and 18 are descriptive only.

\subsection{Human, paraphrase, and language checks}

\begin{table}[h]
\centering
\caption{Prompt-index calibration checks. Human results use a disagreement-enriched stress-test sample. Language results test rescoring after re-expression, not generation or execution.}
\label{tab:measurement-checks}
\begin{tabular}{@{}llrrr@{}}
\toprule
Check & Comparison & $n$ & Pearson $r$ & Additional statistic \\
\midrule
Human, full set & First grader vs ensemble & 200 & 0.408 & ICC$(2,1)=0.395$ \\
Human, overlap & First grader vs ensemble & 50 & 0.613 & mean offset $=-0.69$ \\
Human, overlap & Second grader vs ensemble & 50 & 0.921 & mean offset $=-1.35$ \\
Human, overlap & First vs second grader & 50 & 0.561 & ICC$(2,1)=0.554$ \\
Paraphrase & Original vs plain rewrite & 150 & 0.987 & Spearman $\rho=0.963$ \\
Java & Python vs Java composite & 117 & 0.992 & Java ICC $=0.913$ \\
C++ & Python vs C++ composite & 117 & 0.969 & C++ ICC $=0.882$ \\
\bottomrule
\end{tabular}
\end{table}

The paraphrases have median character length 46.3\% of the originals. Their mean composite shift is $-0.115$ (paired $p=0.031$), with 91.3\% within one point and 99.3\% within two. On the same 117 prompts used for language re-expression, Python inter-judge ICC is 0.912. The per-dimension quadratic-weighted $\kappa$ values in Section~\ref{sec:calibration} compare integer human grades with ensemble means rounded to the nearest integer (half to even).

\subsection{Generations with no model response}
\label{app:no-response}

For 157 generations the model API returned nothing at all: 156 from Gemini 3.1 Pro Preview, from a single batch, with deadline, cancellation, and empty-response errors, and one from Gemini 3 Flash. Of these, 152 are in the later frame and 144, all from Gemini 3.1 Pro Preview, are at composite $>14$. The primary analysis scores them like any generation without code (pass rate 0). Table~\ref{tab:no-response} treats them as missing instead: the pooled outcome averages over the models that responded, and model-specific fits drop those rows. The pooled breakpoint does not move, but Gemini 3.1 Pro Preview's unadjusted contrast turns upward under every outcome definition. Its later-frame decline shrinks (under the primary outcome, 0.890 to 0.869 at 14.5) but does not vanish. Kimi K2.5 has 49 generations without code, but each is a real response written in prose rather than code (its responses went through an extra cleaning pass that marks prose-only answers as code-free), so they stay failures. Its later-frame result is unaffected either way.

\begin{table}[h]
\centering
\caption{Main results with the 157 no-response generations scored as failures (primary analysis) and treated as missing. Cells give $\hat{\gamma}$; mean pass at or below / above it.}
\label{tab:no-response}
\small
\setlength{\tabcolsep}{4pt}
\begin{tabular}{@{}llccc@{}}
\toprule
 & No response as & Independent audit & Submission audit & Raw harness \\
\midrule
Pooled & failures & 14.0; 0.834 / 0.901 & 13.75; 0.799 / 0.876 & 11.25; 0.760 / 0.833 \\
 & missing & 14.0; 0.834 / 0.907 & 13.75; 0.799 / 0.881 & 11.25; 0.760 / 0.836 \\
Gemini 3.1 Pro & failures & 10.75; 0.872 / 0.793 & 11.25; 0.818 / 0.748 & 11.25; 0.784 / 0.738 \\
 & missing & 14.25; 0.859 / 0.876 & 14.25; 0.808 / 0.831 & 14.25; 0.780 / 0.823 \\
\midrule
Models lower above $\hat{\gamma}$ & failures & 2 & 5 & 1 \\
 & missing & 1 (Kimi K2.5) & 4 & 0 \\
\bottomrule
\end{tabular}
\end{table}

\clearpage
\section{Outcome audit}
\label{app:outcome-audit}

\textbf{Why the outcome is audited.} A unit-test verdict can be wrong in both directions: the supplied tests call specific function names and expect exact output formats, so valid code can fail them, while weak tests can pass defective code. The submitted version of this paper asked \texttt{o4-mini} whether each harness verdict was right on the merits, given the task, code, tests, and harness pass rate, and to treat function-name mismatches, script-style input and output, and formatting differences as correct; ``correct'' set the pass rate to 1.0, ``incorrect'' to 0.0, and ``uncertain'' kept the harness value. That audit covered only the 2{,}246 earlier-frame prompts, where it gave a definite verdict for 46{,}706 of 47{,}166 generations (15{,}725 rose to 1.0 and 8{,}123 fell to 0.0) and ``uncertain'' for 92 more. The other 368 received no verdict because the audit call failed (238 unparseable replies, 121 rate-limit errors, and 9 other errors) and keep their raw harness value. Its outcome definition therefore coincided with the construction-frame indicator, and \texttt{o4-mini} is also a rubric judge. For this version, one auditor from a vendor used nowhere else in the study (not a rubric judge, an evaluated model, or the paraphraser) reviewed every generation with code, using the same system prompt, inputs, and verdict mapping.

\textbf{Selection protocol.} A protocol written before any auditor was scored fixed a known-answer set and a selection rule. The set draws 150 benchmark prompts, stratified across display bins, whose reference solution passes every unit test, and builds three cases from each, all executed in the benchmark harness inside a network-isolated container: the normalized reference (clean; correct), a copy with the test-called functions renamed so that the harness fails it (cosmetic; correct), and the most subtle single AST mutation that the tests detect (bug; incorrect). An auditor is eligible if at most 2\% of its responses are unparseable or failed, it judges at least 95\% of clean cases correct, and it rescues at least 90\% of cosmetic cases; among eligible auditors, the lowest wrong-rescue rate on bugs wins.

\textbf{How selection actually went.} Five dated amendments are recorded in the repository, which also collects every departure from the protocol. (i) Before any pilot case was scored, a smoke test showed one candidate misreading the closing format reminder, so the reminder was reworded identically for all auditors and the token cap raised to 16{,}000. (ii) Neither Azure-deployed candidate was eligible (Table~\ref{tab:auditor-pilot}), so we extended the pool to hosted models from unused vendors, to be run in a listed order, and required that a selected auditor also meet the rule on a fresh, disjoint known-answer set at the production endpoint. Only the first two listed candidates were run, and both failed; we then added MiMo-V2.6-Pro, which was not on the list, and the other three listed candidates were never run. (iii) MiMo-V2.6-Pro had the highest overall accuracy and the best cosmetic rescue but judged 140 of 150 clean references correct against the required 143. No candidate met the rule, and we adopted it by author decision before its fresh-set result existed, which made that confirmation report-only. It then met every threshold on the fresh set, with clean accuracy exactly at the minimum. (iv) Generations with no code are marked incorrect by rule without a request, a behavior of the audit client that the protocol did not state. Amendment~5, added after production when a review of the pipeline found that the parser could read a verdict from a response cut off mid-reasoning, stops counting responses cut off by the token limit or a content filter: 9 of MiMo-V2.6-Pro's production verdicts now keep the harness value, and among the pilot rows only Phi-4-reasoning's changes.

\textbf{What the known-answer rates mean.} The rates measure agreement with labels that come from the harness and from how the cases were built, not from human review, and those labels are noisy in both directions. Thirty-five of the 150 clean references in set A are rejected by at least one auditor in Table~\ref{tab:auditor-pilot}, some for violating an explicit task instruction the tests do not check, and at least one other auditor also rejects 7 of the 9 references that MiMo-V2.6-Pro rejects. Conversely, some clean references are defective in ways the tests miss (for example, a JSON parser that does not decode string escapes), and an auditor that accepts them is credited as correct. No human review of the disputed references was performed. The cases are also synthetic renamings and single mutations, which may be easier than real model failures, and an auditor that rescues renamed or reformatted code can also rescue code the tests correctly reject (1.3\% and 0.7\% of bugs on the two sets).

\begin{table}[h]
\centering
\caption{Auditors on the 450-case known-answer sets. Error is the share of unparseable, failed, or cut-off responses (amendment 5). Clean and cosmetic are shares judged correct and wrong rescue is the share of bugs judged correct. An errored case is a miss in every column, so an errored bug case is neither a wrong rescue nor a detection; Overall averages clean, cosmetic, and bug detection (bugs judged incorrect), which is why it can fall below what the other columns imply. \texttt{o4-mini}, run with the submission audit's prompt and settings, is a diagnostic of that audit, not a candidate. The last auditor row is the fresh confirmation set.}
\label{tab:auditor-pilot}
\small
\begin{tabular}{@{}llrrrrr@{}}
\toprule
Auditor & Endpoint & Error & Clean & Cosmetic & Wrong rescue & Overall \\
\midrule
Phi-4-reasoning & Azure & 0.009 & 0.887 & 0.467 & 0.053 & 0.764 \\
MAI-Thinking-1 & Azure & 0.000 & 0.953 & 0.880 & 0.033 & 0.933 \\
Nemotron-3-Ultra & OpenRouter, fp8 & 0.004 & 0.860 & 0.787 & 0.000 & 0.878 \\
Laguna-S-2.1 & OpenRouter, fp4 & 0.047 & 0.887 & 0.747 & 0.033 & 0.864 \\
MiMo-V2.6-Pro & OpenRouter, bf16 & 0.002 & 0.933 & 0.947 & 0.013 & 0.956 \\
\texttt{o4-mini} (submission) & Azure & 0.018 & 0.880 & 0.820 & 0.000 & 0.891 \\
\midrule
MiMo-V2.6-Pro, fresh set & OpenRouter, bf16 & 0.000 & 0.953 & 0.973 & 0.007 & 0.973 \\
\midrule
Selection rule & & $\leq0.02$ & $\geq0.95$ & $\geq0.90$ & lowest & \\
\bottomrule
\end{tabular}
\end{table}

\textbf{Production audit.} Pinned to a single bf16 provider endpoint with streamed responses, MiMo-V2.6-Pro reviewed the 104{,}792 generations with code; the other 208 are incorrect by rule. Counting only complete responses, it returned 88{,}999 correct and 14{,}656 incorrect verdicts (14{,}864 with the rule cases), 177 uncertain, and 960 unparseable or cut off. Of 65{,}350 generations that pass every test, it confirms 64{,}522 and overturns 812; of 12{,}191 that pass none, it judges 7{,}014 correct; and of 27{,}459 partial passes, it sets 17{,}463 to 1.0 and 9{,}227 to 0.0. The auditor's stated reasons for the 7{,}014 rescues of generations that pass no test are mostly test-harness artifacts: 77\% cite a harness, fixture, or test-side problem, for example a function or class name that differs from the one the tests call (about 26\%), a script-style \texttt{input()} or \texttt{main()} call that blocks on import (about 17\%), or an output format or type difference (about 15\%). These are the auditor's own explanations, classified by keyword; the known-answer sets in this appendix measure how often such a rescue is wrong. Mean pass is 0.851, against 0.792 on raw harness values and 0.820 under the submission audit; it is 0.795 in the earlier frame (harness 0.683) and 0.897 in the later frame (harness 0.880). The two Azure candidates also audited a seeded 5\% sample (5{,}250 generations). Excluding the ten rule-based cases and counting only complete responses, MAI-Thinking-1 agrees with MiMo-V2.6-Pro on 91.1\% of the 5{,}178 generations where both return correct or incorrect (Cohen $\kappa=0.631$), and Phi-4-reasoning on 81.2\% of 5{,}029 ($\kappa=0.425$).

\begin{table}[h]
\centering
\caption{Main breakpoint analyses under three outcome definitions: the independent audit (primary), the submission audit (\texttt{o4-mini}, earlier frame only), and raw unit-test harness outcomes. $\Delta$ is the change in percentage points across the selected break: covariate-adjusted for the fixed-effect rows, raw otherwise. Every search rejects the no-break null with zero of 300 wild-bootstrap exceedances, except within the earlier frame ($^\dagger$; $p_{\mathrm{MC}}$ from 0.14 to 0.19). The last row counts models whose mean pass is lower above their own breakpoint, with the 157 no-response generations treated as missing in parentheses.}
\label{tab:outcome-sensitivity}
\small
\begin{tabular}{@{}lrrrrrr@{}}
\toprule
& \multicolumn{2}{c}{Independent audit} & \multicolumn{2}{c}{Submission audit} & \multicolumn{2}{c}{Raw harness} \\
\cmidrule(lr){2-3}\cmidrule(lr){4-5}\cmidrule(l){6-7}
Specification & $\hat{\gamma}$ & $\Delta$ & $\hat{\gamma}$ & $\Delta$ & $\hat{\gamma}$ & $\Delta$ \\
\midrule
Unadjusted & 14.00 & $+6.8$ & 13.75 & $+7.6$ & 11.25 & $+7.4$ \\
Task-type fixed effects & 14.00 & $+0.4$ & 13.75 & $+1.6$ & 11.25 & $+2.0$ \\
Frame fixed effect & 14.25 & $+1.8$ & 14.00 & $+0.8$ & 13.75 & $+0.2$ \\
Task effects differing by side & 11.50 & $+2.1$ & 10.75 & $+2.1$ & 9.50 & $+2.1$ \\
Frame effect differing by side & 14.25 & $+6.8$ & 8.50 & $-3.5$ & 8.50 & $-0.3$ \\
Earlier frame only$^\dagger$ & 9.50 & $-13.0$ & 7.75 & $-15.8$ & 9.50 & $-13.2$ \\
Later frame only & 14.25 & $+1.8$ & 14.25 & $+0.9$ & 14.25 & $+0.9$ \\
Kimi K2.5, later frame & 14.25 & $-4.0$ & 14.25 & $-4.8$ & 14.25 & $-4.8$ \\
\midrule
Models with lower pass above $\hat{\gamma}$ & \multicolumn{2}{c}{2 (1)} & \multicolumn{2}{c}{5 (4)} & \multicolumn{2}{c}{1 (0)} \\
\bottomrule
\end{tabular}
\end{table}

\textbf{Sensitivity to the outcome definition.} Table~\ref{tab:outcome-sensitivity} repeats the main analyses under the three definitions. Several results hold under all three: with additive fixed effects the task-adjusted jump stays small (at most 2.0 points) and the frame-controlled break stays near 14 (13.75 to 14.25); the later frame breaks at 14.25 while the earlier frame shows no significant break; and Kimi K2.5's later-frame decline holds, as does its frame-controlled break at 14.25 (sup-Wald 45.5, 43.6, and 41.9 with an additive frame effect). Three things move: the unadjusted pooled breakpoint (14.0, 13.75, 11.25); the by-side specifications, whose frame-controlled break falls to 8.50 under the two older outcomes; and the set of downward models, which is Kimi K2.5 and Gemini 3.1 Pro Preview under the independent audit, Gemini alone on raw harness outcomes, and those two plus Grok-3, GPT-5.4, and Qwen 3.6 Plus under the submission audit. The extension and fixed-version runs saved no code and the repeated-sampling generations were not audited, so those checks use raw harness outcomes throughout. The protocol and amendments are in \texttt{docs/independent\_audit\_protocol.md}, the audit code in \texttt{src/audit/}, and the reanalysis for any outcome definition in \texttt{scripts/camera\_ready/}.

\section{Dataset, code, and reproducibility details}
\label{app:artifact}

\textbf{Dataset.} The release at \dataurl{} contains the 5{,}000 locked prompts with source identifiers and construction-frame labels; the 19{,}997 per-judge rubric score rows and ensemble aggregates; all 105{,}000 generations with Lizard output complexity, three outcome columns (raw harness, the independent audit, and the submission audit), the auditor's verdicts, a no-response flag, and recorded generation settings; the known-answer audit sets with every auditor's responses and the production verdicts of all three auditors; the repeated-sampling subset; the human calibration grades; and the 365-prompt audit-clean extension. It is documented with a dataset card and Croissant metadata including the Responsible-AI fields. Its \texttt{models} table lists each model's route, served version, quantization, and documented settings, and the three generation system-prompt variants are included. The five locally served models are GPT-OSS-20B, Qwen 3.5-9B, Ministral-3-14B-reasoning, and Devstral Small 2505 (4-bit GGUF) and GLM 4.7-flash (8-bit).

\textbf{Code.} The repository (\codeurl{}) contains the six-dimension rubric prompt, model metadata, the outcome-audit code and protocol, and analysis scripts for the breakpoint, control, construction-frame, diagnostic-IV, placebo, and figure pipelines, together with locked summary outputs and the source tables for every figure. \texttt{scripts/camera\_ready/run\_reanalysis.sh} regenerates every outcome-dependent result for a chosen outcome definition. Re-running model inference or the audit requires provider access; the human-calibration, paraphrase, cross-language, and repeated-sampling statistics can be recomputed from the released tables.

\textbf{Existing assets and licenses.} The source task pool is OpenCodeInstruct under CC BY 4.0~\cite{opencodeinst}, and the derived benchmark is released under CC BY 4.0, with model outputs remaining subject to their providers' terms. 
Generated-output cyclomatic complexity is computed with Lizard 1.21.0 under the MIT License~\cite{lizard}; diagnostic IV estimation uses \texttt{linearmodels} 7.0 under the NCSA License~\cite{linearmodels}. The code repository is released under the MIT License.

\textbf{Compute.} The main compute cost is inference: four rubric judges score 5{,}000 prompts each, and 21 evaluated models each generate one answer for every prompt. Open-weight or locally served models are run one model at a time on A100-class GPU workers; serverless/API models are run through provider batch or hosted inference endpoints. The outcome audit sends one request per generation with code to a hosted auditor (104{,}792 production requests before retries, plus 450 known-answer cases per auditor per set and 5{,}240 agreement-sample requests per second auditor). Unit-test execution, Lizard scoring, and econometric analysis run on CPU workers. Exact wall times, worker memory, and total compute were not logged consistently across providers, so we do not claim a complete compute accounting.

\textbf{Execution safeguards.} Generated code is executed only by the benchmark unit-test harness, one test assertion per fresh Python process in isolated mode with a 5-second timeout. The harness itself does not isolate the file system or the network; the audit's known-answer variants ran in the same harness inside a container without network access, and the auditor reads code but never executes it. The artifact does not release a trained model or a public code-generation service.

\textbf{LLM usage.} LLMs are part of the core methodology as rubric judges, as evaluated code-generation models, and as the outcome auditor. LLM-assisted drafting, editing, and independent auditing of the analysis code were also used under author supervision; all scientific claims, analysis choices, and final text remain the authors' responsibility.

\newpage
\section*{NeurIPS Paper Checklist}

\begin{enumerate}

\item {\bf Claims}
    \item[] Question: Do the main claims made in the abstract and introduction accurately reflect the paper's contributions and scope?
    \item[] Answer: \answerYes{}
    \item[] Justification: The abstract and Introduction present the work as prompt-side measurement and model-specific, benchmark-conditional breakpoint estimation, not causal identification or a universal cutoff. Claims are scoped to the 5,000-prompt selection-stratified benchmark and reported 21-model panel, with construction-frame, task-type, pooling, outcome-definition, human-calibration, model-exclusion, and extreme-tail limitations stated directly.

\item {\bf Limitations}
    \item[] Question: Does the paper discuss the limitations of the work performed by the authors?
    \item[] Answer: \answerYes{}
    \item[] Justification: The paper contains a dedicated Limitations section covering the constructed sampling frame, source and task composition, scorer and human-rater disagreement, pooling sensitivity, dependence on one LLM outcome auditor, sparse support above bin 16, Python-only execution, failed IV restrictions, post hoc robustness checks, and cross-sectional scope.
    \item[] Guidelines:
    \begin{itemize}
        \item The answer NA means that the paper has no limitation while the answer No means that the paper has limitations, but those are not discussed in the paper.
        \item The authors are encouraged to create a separate "Limitations" section in their paper.
        \item The paper should point out any strong assumptions and how robust the results are to violations of these assumptions. The authors should reflect on how these assumptions might be violated in practice and what the implications would be.
        \item The authors should reflect on the scope of the claims made. In general, empirical results often depend on implicit assumptions, which should be articulated.
        \item The authors should reflect on the factors that influence the performance of the approach.
        \item The authors should discuss the computational efficiency of the proposed algorithms and how they scale with dataset size.
    \end{itemize}

\item {\bf Theory assumptions and proofs}
    \item[] Question: For each theoretical result, does the paper provide the full set of assumptions and a complete (and correct) proof?
    \item[] Answer: \answerNA{}
    \item[] Justification: The paper uses standard breakpoint and IV diagnostic estimators but does not introduce new theorems or proofs. The candidate IV restrictions reject, and the manuscript explicitly makes no causal interpretation.

\item {\bf Experimental result reproducibility}
    \item[] Question: Does the paper fully disclose all the information needed to reproduce the main experimental results of the paper to the extent that it affects the main claims and/or conclusions of the paper (regardless of whether the code and data are provided or not)?
    \item[] Answer: \answerYes{}
    \item[] Justification: The Methodology distinguishes the preliminary single-rater sampling score from the four-judge analysis index and specifies the source scan, candidate pooling, exact sampling bands, scoring rubric, evaluated panel, unit-test outcome and its independent audit (with the auditor-selection protocol and known-answer validation), Lizard complete-case definition, breakpoint search, construction-frame and task controls, diagnostic IV specification, and robustness designs. The appendix describes the public dataset release, the code repository, and the source tables for every figure.
    \item[] Guidelines:
    \begin{itemize}
        \item The answer NA means that the paper does not include experiments.
        \item If the paper includes experiments, a No answer to this question will not be perceived well by the reviewers.
        \item If the contribution is a dataset and/or model, the authors should describe the steps taken to make their results reproducible or verifiable.
        \item Depending on the contribution, reproducibility can be supported by code, data, detailed protocols, hyperparameters, random seeds, or sufficient model and system descriptions.
        \item In the case of closed-source models, access to the model may be limited, but there should be some path to reproducing or verifying the results.
    \end{itemize}

\item {\bf Open access to data and code}
    \item[] Question: Does the paper provide open access to the data and code, with sufficient instructions to faithfully reproduce the main experimental results, as described in supplemental material?
    \item[] Answer: \answerYes{}
    \item[] Justification: The benchmark prompts, per-judge and ensemble rubric scores, all 105{,}000 generations with unit-test outcomes, audit verdicts, and Lizard complexity, the audit's known-answer sets, human calibration grades, and the audit-clean extension are publicly released on Hugging Face (\dataurl{}) under CC BY 4.0, with Croissant metadata; the analysis code, rubric prompt, and figure source tables are public on GitHub (Section~\ref{app:artifact}). Re-running model inference additionally requires provider access.
    \item[] Guidelines:
    \begin{itemize}
        \item The answer NA means that the paper does not include experiments requiring code.
        \item Providing as much information as possible in supplemental material is recommended, but including URLs to data and code is permitted.
        \item While open release is encouraged, authors should also preserve anonymity during double-blind review.
        \item Instructions should contain the environment and commands needed to reproduce the results where applicable.
    \end{itemize}

\item {\bf Experimental setting/details}
    \item[] Question: Does the paper specify all the training and test details (e.g., data splits, hyperparameters, how they were chosen, type of optimizer) necessary to understand the results?
    \item[] Answer: \answerYes{}
    \item[] Justification: The paper does not train a new model. It specifies the preliminary-score sampling design, four-judge analysis index, model panel, primary one-generation setting, repeated-sampling subset and temperature, audit-clean extension, unit-test execution, outcome audit, Lizard measurement, and statistical estimators. Main-run generation settings (routes, token limits, system-prompt variants) varied by model and frame and are reported in the released models table and per-row fields where recorded; the paper states that the two frames were generated in separate runs.

\item {\bf Experiment statistical significance}
    \item[] Question: Does the paper report error bars suitably and correctly defined or other appropriate information about the statistical significance of the experiments?
    \item[] Answer: \answerYes{}
    \item[] Justification: The Results and appendix report pairs-bootstrap intervals, sup-Wald statistics, bootstrap and placebo exceedance counts with finite Monte Carlo corrections, Welch tests, pooled and model-specific construction-frame sensitivities, task-type sensitivity, and prompt-level normal-approximation intervals for the tail extension. The output-CC caption labels its band as descriptive generation-row uncertainty and states that it does not adjust for repeated prompts or models.
    \item[] Guidelines:
    \begin{itemize}
        \item The answer NA means that the paper does not include experiments.
        \item The authors should answer Yes if the results are accompanied by error bars, confidence intervals, or statistical significance tests for the experiments that support the main claims.
        \item The factors of variability that the error bars are capturing should be clearly stated.
        \item The method for calculating error bars should be explained.
    \end{itemize}

\item {\bf Experiments compute resources}
    \item[] Question: For each experiment, does the paper provide sufficient information on the computer resources (type of compute workers, memory, time of execution) needed to reproduce the experiments?
    \item[] Answer: \answerNo{}
    \item[] Justification: The appendix identifies serverless/API inference, one-model-at-a-time A100-class GPU workers, and CPU execution for tests, Lizard scoring, and analysis. Exact worker memory, wall times, per-run compute, and total compute were not logged consistently across providers, so the disclosure is incomplete.
    \item[] Guidelines:
    \begin{itemize}
        \item The answer NA means that the paper does not include experiments.
        \item The paper should indicate the type of compute workers CPU or GPU, internal cluster, or cloud provider, including relevant memory and storage.
        \item The paper should provide the amount of compute required for each of the individual experimental runs as well as estimate the total compute.
        \item The paper should disclose whether the full research project required more compute than the experiments reported in the paper.
    \end{itemize}

\item {\bf Code of ethics}
    \item[] Question: Does the research conducted in the paper conform, in every respect, with the NeurIPS Code of Ethics?
    \item[] Answer: \answerYes{}
    \item[] Justification: The work evaluates code-generation reliability using programming prompts and generated code, does not collect personal data or target protected populations, and does not deploy a model. The two authors supplied blind rubric ratings as methodological calibration. The paper reports limitations, artifact access, and execution safeguards.
    \item[] Guidelines:
    \begin{itemize}
        \item The answer NA means that the authors have not reviewed the NeurIPS Code of Ethics.
        \item If the authors answer No, they should explain the special circumstances that require a deviation from the Code of Ethics.
        \item The authors should make sure to preserve anonymity where applicable.
    \end{itemize}

\item {\bf Broader impacts}
    \item[] Question: Does the paper discuss both potential positive societal impacts and negative societal impacts of the work performed?
    \item[] Answer: \answerYes{}
    \item[] Justification: The Introduction, Discussion, Limitations, and Conclusion discuss the value of pre-generation diagnostics and the risks of treating an observational breakpoint as a universal cutoff, causal effect, or deployment rule. Scorer bias, human disagreement, and sparse-tail uncertainty are explicit.
    \item[] Guidelines:
    \begin{itemize}
        \item The answer NA means that there is no societal impact of the work performed.
        \item If the authors answer NA or No, they should explain why their work has no societal impact or why the impact discussion is not applicable.
        \item If there are negative societal impacts, the authors should discuss possible mitigation strategies.
    \end{itemize}

\item {\bf Safeguards}
    \item[] Question: Does the paper describe safeguards that have been put in place for responsible release of data or models that have a high risk for misuse?
    \item[] Answer: \answerNA{}
    \item[] Justification: The paper does not release a trained model or public code-generation service. The appendix nevertheless states the execution safeguards actually used for generated code (one assertion per fresh isolated-mode Python process with a timeout; network-isolated containers for the audit's known-answer runs) and notes that the harness itself does not isolate the file system or network.
    \item[] Guidelines:
    \begin{itemize}
        \item The answer NA means that the paper poses no such risks.
        \item Released models that have a high risk for misuse should describe usage guidelines, restrictions, or safety filters.
        \item Datasets scraped from the Internet could pose safety risks; authors should describe how they avoided releasing unsafe content.
        \item We recognize that providing effective safeguards is challenging, but authors are encouraged to make a best faith effort.
    \end{itemize}

\item {\bf Licenses for existing assets}
    \item[] Question: Are the creators or original owners of assets (e.g., code, data, models), used in the paper, properly credited and are the license and terms of use explicitly mentioned and properly respected?
    \item[] Answer: \answerYes{}
    \item[] Justification: The paper cites OpenCodeInstruct, Lizard, and linearmodels, and names every evaluated model, judge, and auditor with its version; the released models table records each one's developer, access route, and served model string. The appendix names CC BY 4.0 for OpenCodeInstruct and for the derived benchmark, the MIT License for Lizard 1.21.0, and the NCSA License for linearmodels 7.0.
    \item[] Guidelines:
    \begin{itemize}
        \item The answer NA means that the paper does not use existing assets.
        \item The authors should cite the original paper that produced the code package or dataset.
        \item The authors should state which version of the asset is used and, if possible, include a URL.
        \item The name of the license should be included for each asset.
        \item For existing datasets that are re-packaged, both the original license and the license of the derived asset should be provided.
    \end{itemize}

\item {\bf New assets}
    \item[] Question: Are new assets introduced in the paper well documented and is the documentation provided alongside the assets?
    \item[] Answer: \answerYes{}
    \item[] Justification: The new assets are the released benchmark dataset (\dataurl{}), documented with a dataset card and Croissant metadata including Responsible-AI fields, and the analysis code repository (\codeurl{}) with its rubric prompt, locked result summaries, and figure source tables (Section~\ref{app:artifact}).
    \item[] Guidelines:
    \begin{itemize}
        \item The answer NA means that the paper does not release new assets.
        \item Researchers should communicate the details of the dataset/code/model as part of their submissions via structured templates.
        \item The paper should discuss whether and how consent was obtained from people whose asset is used.
        \item At submission time, remember to anonymize your assets if applicable.
    \end{itemize}

\item {\bf Crowdsourcing and research with human subjects}
    \item[] Question: For crowdsourcing experiments and research with human subjects, does the paper include the full text of instructions given to participants and screenshots, if applicable, as well as details about compensation (if any)?
    \item[] Answer: \answerNA{}
    \item[] Justification: No external participants or crowd workers were recruited. The two authors produced blind rubric ratings as part of conducting the study; no personal or sensitive data were collected, and no participant compensation applied.
    \item[] Guidelines:
    \begin{itemize}
        \item The answer NA means that the paper does not involve crowdsourcing nor research with human subjects.
        \item Including this information in the supplemental material is fine, but if the main contribution involves human subjects, then as much detail as possible should be included in the main paper.
        \item According to the NeurIPS Code of Ethics, workers involved in data collection, curation, or other labor should be paid at least the minimum wage in the country of the data collector.
    \end{itemize}

\item {\bf Institutional review board (IRB) approvals or equivalent for research with human subjects}
    \item[] Question: Does the paper describe potential risks incurred by study participants, whether such risks were disclosed to the subjects, and whether Institutional Review Board (IRB) approvals (or an equivalent approval/review based on the requirements of your country or institution) were obtained?
    \item[] Answer: \answerNA{}
    \item[] Justification: No external human subjects or crowd workers were recruited. The only human ratings were research annotations produced by the two authors, with no personal or sensitive data collected.
    \item[] Guidelines:
    \begin{itemize}
        \item The answer NA means that the paper does not involve crowdsourcing nor research with human subjects.
        \item Depending on the country in which research is conducted, IRB approval or equivalent may be required for any human subjects research.
        \item We recognize that the procedures for this may vary significantly between institutions and locations, and we expect authors to adhere to the NeurIPS Code of Ethics and the guidelines for their institution.
        \item For initial submissions, do not include any information that would break anonymity, such as the institution conducting the review.
    \end{itemize}

\item {\bf Declaration of LLM usage}
    \item[] Question: Does the paper describe the usage of LLMs if it is an important, original, or non-standard component of the core methods in this research? Note that if the LLM is used only for writing, editing, or formatting purposes and does not impact the core methodology, scientific rigor, or originality of the research, declaration is not required.
    \item[] Answer: \answerYes{}
    \item[] Justification: LLM usage is central: out-of-panel LLMs score prompts, a 21-model panel is evaluated, an independent LLM auditor defines the primary outcome, and LLMs support paraphrase and task-label robustness checks. The appendix also declares LLM-assisted drafting, editing, and code auditing under author supervision.
    \item[] Guidelines:
    \begin{itemize}
        \item The answer NA means that the core method development in this research does not involve LLMs as any important, original, or non-standard components.
        \item Please refer to the NeurIPS LLM policy in the handbook for what should or should not be described.
    \end{itemize}

\end{enumerate}

\end{document}